\pdfoutput=1
\documentclass[sigconf, natbib=true]{acmart}

\usepackage{graphicx}
\usepackage{amsthm,amsmath}
\usepackage{booktabs}
\usepackage{subfigure}
\usepackage{color}
\usepackage{mathrsfs}
\usepackage{bm}
\usepackage{multirow}
\usepackage{tabularx}
\usepackage[noend]{algpseudocode}
\usepackage{array}
\usepackage{enumitem}
\usepackage{balance}
\usepackage{pdftexcmds}
\usepackage{catchfile}
\usepackage{ifluatex}
\usepackage{ifplatform}
\usepackage{caption}
\usepackage{color}
\usepackage{bbm}
\usepackage[normalem]{ulem}

\newcolumntype{L}[1]{>{\raggedright\let\newline\\\arraybackslash\hspace{0pt}}m{#1}}
\newcolumntype{C}[1]{>{\centering\let\newline  \\\arraybackslash\hspace{0pt}}m{#1}}
\newcolumntype{R}[1]{>{\raggedleft\let\newline \\\arraybackslash\hspace{0pt}}m{#1}}

\usepackage{booktabs}
\usepackage{pdfpages}

\usepackage{algorithm} 
\usepackage{algorithmicx}
\usepackage{algpseudocode}
\usepackage{url}

\def\bA{\textbf{A}}

\def\bS{\textbf{S}}
\def\bH{\textbf{H}}
\def\bW{\textbf{W}}
\def\bal{\bm{\alpha}}
\DeclareMathOperator*{\argmin}{arg\,min}
\def\bh{\textbf{h}}
\def\bz{\textbf{z}}

\AtBeginDocument{%
  \providecommand\BibTeX{{%
    \normalfont B\kern-0.5em{\scshape i\kern-0.25em b}\kern-0.8em\TeX}}}

\copyrightyear{2021}
\acmYear{2021}
\setcopyright{acmcopyright}
\acmConference[CIKM '21] {Proceedings of the 30th ACM International Conference on Information and Knowledge Management}{November 1--5, 2021}{Virtual Event, Australia.}
\acmBooktitle{Proceedings of the 30th ACM Int'l Conf. on Information and Knowledge Management (CIKM '21), November 1--5, 2021, Virtual Event, Australia}
\acmPrice{15.00}
\acmISBN{978-1-4503-8446-9/21/11}
\acmDOI{10.1145/3459637.3482285}

\begin{document}
\fancyhead{}
\title{Pooling Architecture Search for Graph Classification}

%\author{Quanming Yao}
%\email{qyaoaa@tsinghua.edu.cn}
%\affiliation{%
%  \institution{Department of Electronic Engineering, Tsinghua University}
%  \institution{4Paradigm. Inc.}
%  \streetaddress{P.O. Box 1212}
%  \city{Beijing}
%  \state{}
%  \postcode{}
%  \country{China}
%}

%\author{Lanning Wei$^{1,3}$, 
%	Huan Zhao$^1$, 
%	Quanming Yao$^{1,2}$,
%	Zhiqiang He$^{2,4}$}
%\affiliation{
%	\institution{$^3$Institute of Computing Technology, Chinese Academy of Sciences},
%	\institution{$^4$University of Chinese Academy of Sciences},
%	\institution{$^1$4Paradigm Inc., Beijing, China},
%	\institution{$4$Department of Electronic Engineering, Tsinghua University}
%}
%\email{weilanning18z@ict.ac.cn;zhaohuan@4paradigm.com;hezq@lenovo.com}
\author{Lanning Wei$^{1,2,3}$, 
	Huan Zhao$^3$, 
	Quanming Yao$^{3,4}$
	Zhiqiang He$^{1,5}$}
\affiliation{
	\institution{$^1$Institute of Computing Technology, Chinese Academy of Sciences $^2$University of Chinese Academy of Sciences $^3$4Paradigm. Inc., $^4$Department of Electronic Engineering, Tsinghua University, $^5$Lenovo}
%	\institution{$^1$4Paradigm Inc.}
	\city{Beijing}
	\country{China}
}
\email{weilanning18z@ict.ac.cn; zhaohuan@4paradigm.com; qyaoaa@tsinghua.edu.cn; hezq@lenovo.com}

%\author{Lars Th{\o}rv{\"a}ld}
%\affiliation{%
%  \institution{The Th{\o}rv{\"a}ld Group}
%  \streetaddress{1 Th{\o}rv{\"a}ld Circle}
%\email{larst@affiliation.org}


%\renewcommand{\shortauthors}{Wei and Zhao, et al.}

\begin{abstract}
Graph classification is an important problem with applications across many domains, like chemistry and bioinformatics, 
for which graph neural networks (GNNs) have been state-of-the-art (SOTA) methods.
GNNs are designed to learn node-level representation based on neighborhood aggregation schemes, and 
to obtain graph-level representation, pooling methods are applied after the aggregation operation in existing GNN models to generate coarse-grained graphs.
However, due to highly diverse applications of graph classification, and the performance of existing pooling methods vary on different graphs. 
In other words, it is a challenging problem to design a universal pooling architecture to perform well in most cases, 
leading to a demand for data-specific pooling methods in real-world applications.
To address this problem, we propose to use neural architecture search (NAS) to search for adaptive pooling architectures for graph classification. 
Firstly we designed a unified framework consisting of four modules: Aggregation, Pooling, Readout, and Merge, which can cover existing human-designed pooling methods for graph classification.
Based on this framework, a novel search space is designed by incorporating popular operations in human-designed architectures. 
Then to enable efficient search, a coarsening strategy is proposed to continuously relax the search space, thus a differentiable search method can be adopted.
Extensive experiments on six real-world datasets from three domains are conducted, and the results demonstrate the effectiveness and efficiency of the proposed framework\footnote{Lanning and Huan contribute equally to this work, and Lanning is a research intern in 4Paradigm. Zhiqiang He is the corresponding author. The implementation of PAS is available at: \url{https://github.com/AutoML-Research/PAS}}.
\end{abstract}

%\begin{CCSXML}
%<ccs2012>
%  <concept_id>10010520.10010553.10010562</concept_id>
%  <concept_desc>Computer systems organization~Embedded systems</concept_desc>
%  <concept_significance>500</concept_significance>
%  <concept_id>10010520.10010575.10010755</concept_id>
%  <concept_desc>Computer systems organization~Redundancy</concept_desc>
%  <concept_significance>300</concept_significance>
%  <concept_id>10010520.10010553.10010554</concept_id>
%  <concept_desc>Computer systems organization~Robotics</concept_desc>
%  <concept_significance>100</concept_significance>
%  <concept_id>10003033.10003083.10003095</concept_id>
%  <concept_desc>Networks~Network reliability</concept_desc>
%  <concept_significance>100</concept_significance>
%</ccs2012>
%\end{CCSXML}
%
%\ccsdesc[500]{Computer systems organization~Embedded systems}



\maketitle

\section{Introduction}
\label{sec-intro}
%\wei{1. one sentence to introduce GNN? (start from graph classification. dr.yao)
%	2. the main difficulties to apply GNNs to graph classification (graph-level information).
%	3. how to solve? recent GNN use pooling to capture graph-level information (global+hierarchical)\\
% 	4. challenges: fixed pooling + various task + generalize, various task only mentioned once in here.
% 	5. graphgym experiment to vertify the solution: fixed pooling cannot generalized well. (we need not only data-specific but also adaptive pooling.)\\
% 	6. solutions: NAS, (NAS related work)
% 	7. cannot be directly used for graph classification.\\
% 	8. above challenges: data-specific + adaptive pooling architecture.}
%In recent years, graph neural networks (GNNs) have been the state-of-the-art (SOTA) method for node classification task~\cite{kipf2016semi,hamilton2017inductive,velivckovic2017graph} and link prediction task~\cite{zhang2018link}, which designed to learn node-level representation based on neighborhood aggregation scheme~\cite{gilmer2017neural}. 
%\wei{start from GNNs~\cite{yuan2019structpool,yuan2020xgnn}}\huan{It is better to start from graph classification. Think it a bit more.}
%
%However, graph classification task needs to learn graph-level representations. 
%

In recent years, graph neural networks (GNNs)~\cite{zhang2018end,ying2018hierarchical} have been the state-of-the-art (SOTA) method for graph classification, which is an important problem with applications in various domains, e.g., chemistry~\cite{gilmer2017neural}, bioinformatics~\cite{ying2018hierarchical}, text categorization~\cite{rousseau2015text}, social networks~\cite{xu2018powerful}, and recommendation~\cite{xiao2019beyond,gao2021efficient}.
Most GNNs, 
e.g., GCN~\cite{kipf2016semi}, GraphSAGE~\cite{hamilton2017inductive}, GAT~\cite{velivckovic2017graph} and GIN~\cite{xu2018powerful},
%%\footnote{+qm+ list some famous names in Table~3 here.\weicheck{}}
were designed to learn node-level representation based on neighborhood aggregation schemes~\cite{gilmer2017neural}, i.e., the embedding of a node is updated by iteratively aggregating the embeddings of its neighbors in a graph. 
However, for the graph classification task, we need to generate graph-level representation from the node embeddings given a graph.

To achieve this, 
%\textit{pooling} methods are designed together with neighborhood aggregation operations, 
%which try to generate a fixed-size embedding vector as the graph representation from all node embeddings.
in the literature, 
various \textit{pooling} methods have been proposed 
%to obtain graph-level representation 
based on GNNs.
The simplest method is directly taking the 
%%\footnote{+qm+ you already say ``all node embeddings''.\weicheck{}}
mean or summation of all node embeddings as the graph representation~\cite{xu2018powerful}. 
%%\footnote{+qm+ before stating ``a lack of hierarchical information'',
%	it is better explain what is it and why it is important first. \label{ft:1}\weicheck{}}
However, the \textit{global pooling} methods only use node features and ignore the hierarchical information as they do not exploit the graph compositional nature~\cite{khasahmadi2020memory},
%that might exist in the graph, 
leading to \textit{flat} graph-level representation~\cite{ying2018hierarchical}. 
Then more advanced pooling methods are proposed to preserve the hierarchical information by aggregating messages on coarser and coarser graphs, which are generated by applying a pooling operation to reduce the size of a graph after an aggregation operation in each layer (see Figure~\ref{fig-framework}(a)).
%, to reduce the size of a graph iteratively. 
%Thus hierarchical structural information can be preserved when obtaining the graph-level representation in the final layer. 
These methods are dubbed \textit{hierarchical pooling}, and representative ones are DiffPool~\cite{ying2018hierarchical}, SAGPool~\cite{lee2019self}, ASAP~\cite{ranjan2020asap}, Graph U-Net~\cite{gao2019graph}, and STRUCTPOOL~\cite{yuan2019structpool}, etc., which assume a cluster property underlying the graph and generate coarse graph in each layer corresponding to the cluster.

\begin{figure}[t]
	%	\vspace{-10pt}
	\centering
	\includegraphics[width=0.9\linewidth]{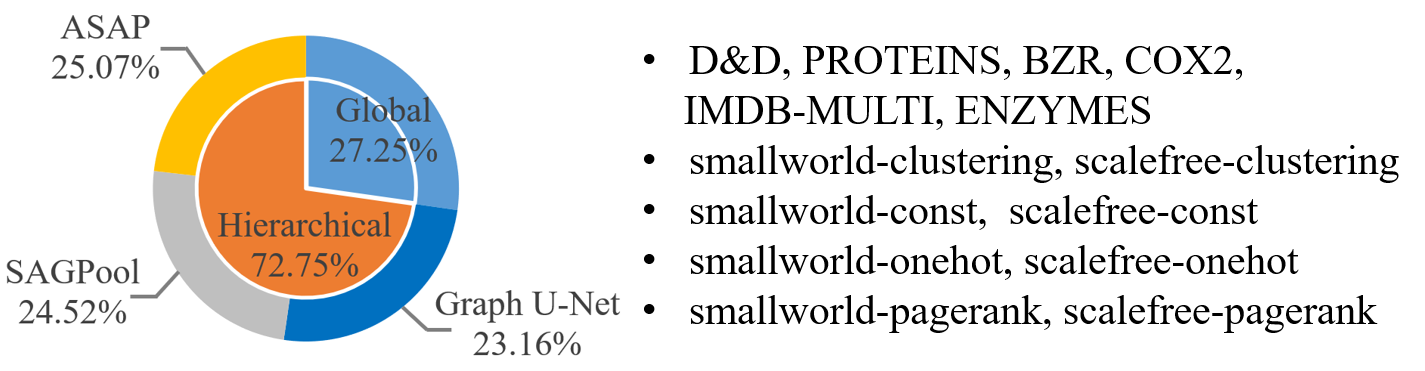}
	\caption{The performance comparisons between four popular pooling methods on 14 graph classification datasets, with 6 real-world and 8 synthetic ones. Left: We sample 420 setups based on GraphGym and shown the percentage distribution of each method winning over the other three. Right: The specific 14 graphs from GraphGym, 6 real-world datasets and 8 synthetic datasets with different structures and features. More details are in Section~\ref{sec-exp-graphgym}.}
	\label{fig-graphgym-winning-ratio}
	%	\vspace{-20pt}
\end{figure}

Despite the success of these pooling methods, in reality, graphs are from highly diverse domains, 
e.g., chemistry~\cite{gilmer2017neural}, bioinformatics~\cite{ying2018hierarchical}, text categorization~\cite{rousseau2015text}, social networks~\cite{xu2018powerful} and recommendation~\cite{xiao2019beyond,gao2021efficient},
leading to a challenging problem that human-designed pooling architectures 
%%\footnote{+qm+ better no say so.
%	the best choices in Figure~1 still are all human designed ones.
%	The problem is that human designed one cannot well adapt to different datasets.\weicheck{}} 
%cannot generalize well in graph classification.
cannot adapt to diverse datasets well.
%%\footnote{+qm+ better show our results instead.\huancheck{The experiments based on GraphGym can be more convincing in my opinion.}}
To verify this problem,  we design an experiment based on a recent GNN benchmark GraphGym~\cite{you2020design} to compare the performance of several representative pooling methods for graph classification.
From Figure \ref{fig-graphgym-winning-ratio}, we can observe that the percentage of each method winning over the other three is very close (around 25\%), which means that no single human-designed pooling architecture can win in all cases (420 setups). Besides, the hierarchical pooling method is not always superior to the global one, which is consistent with the two latest works~\cite{errica2019fair,degio2020rethinking}.
%
%Moreover, two latest works~\cite{errica2019fair,degio2020rethinking} show that the hierarchical pooling scheme is not a must for obtaining high-quality graph representations depending on the graphs.
%and in some cases the aggregation and pooling functions can even be harmful to graph representation learning. 
Therefore, it is very important to design data-specific pooling architectures for the graph classification task.

Vere recently, to obtain data-specific GNN architectures, researchers turn to neural architecture search (NAS) \cite{zoph2016neural,liu2018darts}, e.g., GraphNAS \cite{gao2019graphnas} and Policy-GNN \cite{lai2020policy}.
%%\footnote{+qm+ 
%	this sentence is vague,
%	to make reviewers convinced you need to explain the key difference between node and graph task.
%	this means you need to address footnote~\ref{ft:1} first,
%	and then come back.\huancheck{}}
However, most of existing NAS methods for GNN are focusing on the aggregation functions and only use the global pooling functions on top of the searched architecture when dealing with the graph-level tasks. Thus, they fail to obtain data-specific pooling architectures, either. It is non-trivial to design a NAS method for pooling architecture search since both global and hierarchical pooling methods are effective in different scenarios (Figure \ref{fig-graphgym-winning-ratio}). Any NAS method needs to take into consideration these two pooling paradigms. Moreover, the search efficiency also needs to be considered when applying NAS to designing data-specific pooling architectures.

In this work, to the best of our knowledge, we made the first attempt to address the two aforementioned problems and propose an efficient NAS method to obtain data-specific pooling architectures for graph classification. Firstly, by revisiting various existing human-designed pooling architectures, we propose a unified pooling framework consisting of four key modules for graph classification, which covers both the global and hierarchical pooling methods. Then based on the unified framework, a customized and effective search space is designed. To enable efficient search on top of the search space, a differentiable search algorithm could be adopted, which tends to relax the discrete search space into a continuous one by mixing the output of all candidate operations. However, it is infeasible to directly relax the designed search space as existing differentiable search methods like DARTS \cite{liu2018darts} and SNAS \cite{xie2018snas}, since different candidate pooling operations in the search space generate different coarse graphs consisting of diverse nodes and edges. To address this challenge, we design a coarsening strategy to properly relax the selection of pooling operations in an continuous manner, and then develop a differentiable search method to complete the search process.
Finally, we extract the optimal architecture when the searching process terminates. 
In this way, data-specific architectures are obtained, and the proposed method is dubbed PAS (Pooling Architecture Search).
To demonstrate the effectiveness of PAS, 
we conduct extensive experiments on six real-world datasets from three domains, and experimental results show that the searched architectures can outperform various baselines for graph classification.
%not only outperform various baselines for graph classification, but also are in a moderate size in terms of model parameters. 
Moreover, the search time is reduced by two orders of magnitude compared to RL based methods. 
%%\footnote{+qm+ this example argues that ``search GNN + certain = poor performance''.
%	I do not think such an example is needed.
%	1). you have already argue pooling should be data-dependent in last paragraph;
%	2). what you want to say is the design of the search space is not trivial,
%	and their search method cannot be simply extended.
%	So, it is better summarize technical difficulties in Section~3.1 and 3.2 here,
%	and use them to support what you want to claim.\weicheck{}}

To summarize, the contributions of this work are as follows:
\begin{itemize}[leftmargin=*]
	\item To the best of our knowledge, PAS is the first method to learn data-specific pooling architectures for graph classification. To apply NAS to this task, we propose a unified framework which can cover various human-designed pooling architectures, including global and hierarchical ones.
%	\item We design one unified framework to cover the existing methods, it contains 4 essential modules, Aggregation, Pooling, Readout and Merge, respectively.
	\item Based on the unified framework, we design a customized and effective search space. And to enable differentiable architecture search, a coarsening strategy is designed to relax the search space into a continuous one, thus we develop an efficient search method.
	\item Extensive experiments on six real-world datasets from bioinformatics, chemistry, and social networks show that the searched architectures by PAS outperform various baselines for graph classification and the efficiency of PAS in terms of search cost.
\end{itemize}

\noindent\textbf{Notations.} 
We represent a graph as $G =(\bA, \bH) $ ,where $\textbf{A} \in \mathbb{R}^{N \times N}$ is the adjacency matrix of this graph and $\bH \in \mathbb{R}^{N \times d}$ is the node features.  $N$ is the node number.
$\widetilde{\mathcal{N}}(v) =  \{v\} \cup \{ u | \textbf{A}_{uv} \ne 0 \} $ represents set of the self-contained first-order neigbors of node $v$.
 Give a dataset $\mathcal{D} = \left\{ (G_1, y_1), \cdots, (G_M, y_M) \right\}$, $(G_i, y_i)$ is the $i$-th graph of this dataset. $M$ is the number of total graphs, $y \in \mathcal{Y}$ is the graph label. 
In a $L$-layer GNN, for clear presentation, the input graph is denoted by $G^0 = (\bA^0, \bH^0)$, and the input of $l$-th layer is $G^{l-1} = (\textbf{A}^{l-1}, \textbf{H}^{l-1})$, and the output is $G^{l} = (\textbf{A}^{l}, \textbf{H}^{l})$. 
The features of node $v$ in $l$-th layer are denoted by $\textbf{h}_v^l$.

%\newpage

\section{Related Work}
\label{sec-related}

\subsection{GNN for Graph Classification}

\begin{figure*}[t]
	\centering
	\includegraphics[width=0.85\linewidth]{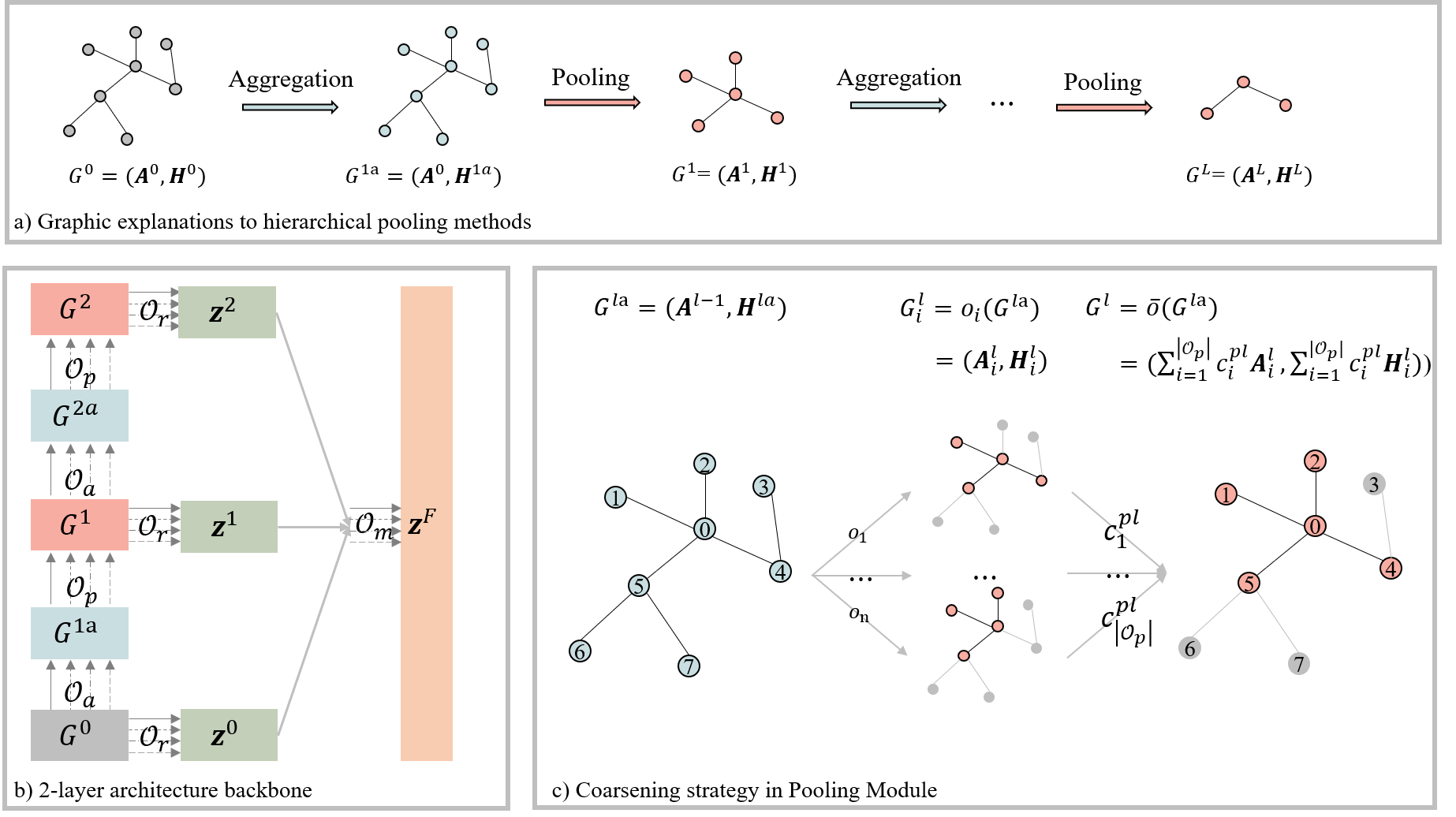}
	\caption{(a) In general, hierarchical methods use one aggregation and one pooling operation in each layer, 
		which is responsible for update node embeddings and generate the coarse graph. When we remove the pooling operations in the intermediate layers and only keep one in the final layer, it leads to global methods.
		(b) We choose a 2-layer supernet as an illustrative example of the unified framework.
%		 and the supernet representing the search space. 
		 Each layer contains 1 Aggregation Module and 1 Pooling Module. 
		Merge Module used to incorporate 3 intermediate graph representations generated by Readout Module. 
		(c) The coarsening strategy we used. For unselected nodes and edges (in grey), 
		we set the features and weights to $0$ so different coarse graph $G_i^{l}$ can be summarized directly.}
	\label{fig-framework}
\end{figure*}

In the literature, existing GNN methods for graph classification can be roughly classified into two groups: global pooling and hierarchical pooling methods. 
Global pooling methods only use one global pooling function behind the final aggregation operation, 
and hierarchial methods use pooling operation after each aggregation in the architecture. 
On one hand, the global pooling methods are straightforward, which add a simple pooling operation, such as global summation of all node embeddings, however, 
as mentioned in~\cite{ying2018hierarchical}, these global pooling methods, e.g., GIN~\cite{xu2018powerful} and DGCNN~\cite{zhang2018end}, learn flat graph embeddings 
%\sout{with different aggregators and global pooling functions}
, which cannot capture the 
potential hierarchical information in real-world graphs.
%%\footnote{+qm+ see footnote~\ref{ft:1}.}
On the other hand, hierarchical pooling methods are proposed to solve this problem by aggregate messages on coarser and coarser graphs, e.g., from $G^0$ to $G^L$ as shown in Figure~\ref{fig-framework}(a). It is achieved by applying a pooling operation to reduce the size of a graph after an aggregation operation in each layer.
%%\footnote{+qm+ do not just list what are these pooling methods.
%	instead, try to explain what they are good / bad at. \label{ft:4}}
For these hierarchical pooling methods, 
SAGPool~~\cite{lee2019self}, 
Graph U-Net~\cite{gao2019graph} and ASAP~~\cite{ranjan2020asap}
sample a set of nodes based on diverse node score functions and form corresponding coarse graphs; 
DiffPool~~\cite{ying2018hierarchical} and 
STRUCTPOOL~\cite{yuan2019structpool}
focus on grouping nodes into clusters with different assignment functions, re-generate the edges among these clusters.
Besides, GMN~~\cite{khasahmadi2020memory} proposes one memory layer to jointly learn node representations and coarsen graph, MinCutPool~\cite{bianchi2020spectral} and EigenPool~\cite{ma2019graph} focus on learning assignment functions in the frequency domain.

However, existing methods use 
%%\footnote{+qm+ the word ``fixed'' is not good.
%	while our operation is searched, the possibility of operation choice is still fixed.
%	it is better say ``adaptive'' or ``data-specific''.\weicheck{}}
predefined pooling operations, 
which are difficult to adapt to various datasets. 
In this paper, by the proposed PAS, we can obtain data-specific pooling architectures for graph classification.

\subsection{Graph Neural Architecture Search}

\label{sec-nas-gnn}
NAS methods were proposed to automatically find SOTA CNN architectures in a pre-defined search space 
%\sout{compared with human-designed ones}, 
and representative methods are~\cite{zoph2016neural,real2017large,pham2018efficient,liu2018darts,xie2018snas,real2019regularized}. Very recently, researchers tried to automatically design GNN architectures by NAS. The majority of these methods focus on design the aggregation layers in GNNs with different search algorithms.  
For example, GraphNAS~\cite{gao2019graphnas}, Auto-GNN~\cite{zhou2019auto}, AutoGM~\cite{yoon2020autonomous}, DSS~\cite{li2021one} and~\cite{peng2019learning} learn to design aggregation layers with diverse dimensions, such as attention function, attention head number, embedding size, etc; SANE~\cite{zhao2021search}, SNAG~\cite{zhao2020simplifying} and AutoGraph~\cite{li2020autograph} provide the extra skip connections learning; GNAS~\cite{cai2021rethinking} and Policy-GNN~\cite{lai2020policy} learn to select the best message passing layers. 
%~\cite{peng2019learning} learns to design aggregations in action recognition. 
Apart from design aggregation layers, RE-MPNN~\cite{jiang2020graph} learns adaptive global pooling functions additionally. However, these methods fail to obtain the data-specific pooling architectures because the pooling operations which are essential to graph classification are not considered.
As to the search algorithm, most of the existing methods use the RL (Reinforcement Learning) and EA(Evolutionary Algorithm) based methods to select architectures from the search space. RL based algorithms, used in ~\cite{gao2019graphnas,zhou2019auto,zhao2020simplifying,lai2020policy}, sample architectures with RNN controller and then updated with policy gradient; EA based algorithms, used in~\cite{li2020autograph,jiang2020graph,peng2019learning}, select parent architecture from the search space and generate new architectures with mutation~\cite{real2017large} and crossover~\cite{real2019regularized}, select parent architecture from the search space and generate new architectures with mutation~\cite{real2017large} and crossover~\cite{real2019regularized}. Bayesian optimization is utilized in AutoGM~\cite{yoon2020autonomous}.
These methods need thousands of evaluations which are computationally expensive, and differentiable search algorithms are proposed to solve the efficiency problem.
They construct an over-parameterized network (supernet) and optimize this supernet with gradient descent due to the continuous relaxation of the search space. Representative methods DARTS~\cite{liu2018darts} and SNAS~\cite{xie2018snas} use the Softmax and the Gumble-Softmax functions as the relaxation function, respectively. The differentiable search algorithms are used in SANE~\cite{zhao2021search}, DSS~\cite{li2021one} and GNAS~\cite{cai2021rethinking}
to relax the aggregation dimensions. However, it is difficult to relax the pooling operations because different candidate pooling operations generate different coarse graphs consisting of diverse nodes and edges. Thus, it is a challenge to design one efficient search algorithm for learning data-specific pooling architectures. 
%More graph neural architectures 

More graph neural architecture search methods can be found in~\cite{zhang2021automated,zhang2020autosf,ding2021diffmg,wang2021explainable,zhang2020interstellar,guan2021autoattend}. Compared with existing methods in 
%%\footnote{+qm+ add a column on ``task'' in this table?
%	e.g., node cls / graph cls?}
Table~\ref{tb-compare-existing}, PAS provides one search space that can cover existing pooling methods and one coarsening strategy to develop an efficient data-specific pooling architecture learning method.

\begin{table}[]
	\centering
	\small
	%\footnotesize
%	\caption{Comparing existing GNN and NAS baselines with PAS. We set the Pooling and Readout Module of node-level methods as ``-'' and the search algorithm of hand-designed methods as ``-''.
%		\texttt{A}: Aggregation,
%		\texttt{P}: Pooling,
%		\texttt{R}: Readout,
%		\texttt{M}: Merge.}}
	\caption{Comparing existing human-designed and NAS based pooling methods with PAS. We set the search algorithm of hand-designed methods as ``-''.
	\texttt{A}: Aggregation,
	\texttt{P}: Pooling,
	\texttt{R}: Readout,
	\texttt{M}: Merge.}
	\setlength\tabcolsep{4pt}
\begin{tabular}{c|c|c|c|c|c|c}
	\hline
	\multirow{2}{*}{}     & \multirow{2}{*}{Methods}             & \multicolumn{4}{c|}{Search Space}                 & Search           \\ \cline{3-7} 
	&                                      & \texttt{A} & \texttt{P} & \texttt{R} & \texttt{M} & Algorithm        \\ \hline
	\multirow{3}{*}{GNNs} & GIN~\cite{xu2018powerful}            & $\surd$    & $\times$   & $\surd$    & $\times$   & -                \\ \cline{2-7} 
	& DiffPool~\cite{ying2018hierarchical} & $\surd$    & $\surd$    & $\surd$    & $\times$   & -                \\ \cline{2-7} 
	& SAGPool~\cite{lee2019self}           & $\surd$    & $\surd$    & $\surd$    & $\surd$    & -                \\ \hline
	\multirow{2}{*}{NAS}  & RE-MPNN~\cite{jiang2020graph}        & $\surd$    & $\times$   & $\surd$    & $\surd$    & EA               \\ \cline{2-7} 
	& PAS (proposed)                       & $\surd$    & $\surd$    & $\surd$    & $\surd$    & Gradient Descent \\ \hline
\end{tabular}
\label{tb-compare-existing}
\end{table}

\section{Method}
\label{sec-framework}

In this section, we elaborate on the proposed PAS, including the unified framework for graph classification, followed by the proposed search space and efficient search algorithm with the proposed coarsening strategy.

\subsection{The Unified Framework}

%The unified framework is illustrated as Figure~\ref{fig-framework}(b),
We define a unified framework that consists of four key modules for learning graph-level representation derived form existing pooling architectures, Aggregation, Pooling, Readout and Merge Module, respectively. In general, one Pooling Module is placed after each Aggregation Module in each layer, and Merge Module is utilized to incorporate the intermediate graph representations produced by Readout Module. 
In Figure~\ref{fig-framework}(b), we use a 2-layer architecture backbone as an illustrative example of the unified framework. With the input Graph $G^0$, Aggregation Module updates node embeddings and produce the graph $G^{1a}=(\bA^0, \bH^{1a})$, Pooling Module generates the coarse graph $G^1=(\bA^1,\bH^1)$ behind. 3 Readout Modules used to capture the graph representations $\bz$ in all layers, and Merge Module generates the final graph representation $\bz^F$. Based on this framework, we can unify most existing pooling methods including global and hierarchical ones. 
%When removing the Pooling Modules in all intermediate layers, 
When we remove the pooling operations in the intermediate layers and only keep one in the final layer, 
we obtain a global pooling method. This design can guarantee the flexibility and expressiveness of the proposed framework. 
In Table \ref{tb-compare-existing}, we can see that representative pooling architectures can be covered by this framework in terms of the four modules.

Based on this unified framework, an effective search space can be naturally designed by including human-designed operations, the details of which are given in Table \ref{tb-search-space}. Then different combinations of these operations can be obtained, leading to data-specific pooling architectures, by any search method.
Specially, to incorporate both the global and hierarchical pooling architectures, we add the $NONE$ operation in the Pooling Module which means no pooling operations are used.
In the NAS literature, the architecture searching task is to solve the following bi-level optimization problem:
\begin{align}
\min\nolimits_{\bal \in \mathcal{A}} & \;
\mathcal{L}_{\text{val}} (\bW^*(\bal), \bal),
\label{eq-nested-nas-opt}
\\\
\text{\;s.t.\;} \bW^*(\bal) 
& = \argmin\nolimits_\bW \mathcal{L}_{\text{train}}(\bW, \bal),
\label{eq-nested-nas-opt:2}
\end{align}
where $\mathcal{A}$ represents the search space, $\bm{\alpha}$ represents one candidate architecture in $\mathcal{A}$, and $\bW$ represents the parameters of a model from the search space, and $\bW^*(\bal)$ represents the corresponding operation parameter after training.
$\mathcal{L}_{\text{train}}$ and $\mathcal{L}_{\text{val}}$ are the training and validation loss, respectively.
%In search process, we update $\bm{\alpha}$ and $\bW$ alternately based on the above continuous relaxation. 
Popular NAS methods use RL \cite{zoph2016neural,gao2019graphnas}, EA \cite{guo2019single}, and differentiable \cite{liu2018darts,xie2018snas,zhao2021search} search algorithms. Due to the efficiency superiority, differentiable methods are more preferable in latest NAS methods. In this work, we also adopt the differentiable search paradigm. However, the challenge is that the proposed search space cannot be directly relaxed continuously, which is a prerequisite step for the differentiable search method, due to the fact that different candidate pooling operations in the search space generate different coarse graphs consisting of diverse nodes and edges. 
To address this challenge, we design a coarsening strategy to properly relax the selection of pooling operations, thus the search space, in an continuous manner. In this way, an efficient search process based on gradient descent is enabled.
In Table \ref{tb-compare-existing}, we compare PAS with existing human-designed and NAS based pooling methods in terms of the four modules (search space) and the search algorithm, and
in the remaining part of this section, we introduce in detail the search space and the differentiable search algorithm.

%With this unified framework, NAS can be employed in learning data-specific pooling architectures. 
%To be specific, we propose one novel search space and one coarsening strategy to address the 2 challenges in applying NAS on this framework. The search space contains a set of operations in each module. Especially, to incorporate the global and hierarchical pooling architectures, we use the $NONE$ operation in the Pooling Module which which means no pooling operations are used. The coarsening strategy is designed to relax the pooling operations and enable the usage of differentiable search algorithm. It leads to a more efficient search process than RL and EA based methods. Compared with existing methods in Table \ref{tb-compare-existing}, PAS provides one unified framework which can cover existing methods and provide an efficient search algorithm.
%With the designed search sapce, especially the $NONE$ operation in pooling module, PAS can cover existing global and hierarchical methods as shown in Table~\ref{tb-compare-existing}.
%Furthermore, we design one coarsening strategy to enable the usage of differentiable search algorithm. 
%with the coarsening strategy, we can use differentiable search algorithm and optimize PAS with gradient descent. 
%Therefore, learning data-specific pooling architectures with NAS is equals to solve the bi-level optimization problem as:

\subsection{The Design of the Search Space}
\label{sec-search-space}

Based on the proposed framework, we design one novel search space with a set of candidate operations as shown in Table~\ref{tb-search-space}.
%As shown in Figure~\ref{fig-framework}(b), each node, which represented as one rectangle, denotes the intermediate representations and each edge, which represented as the dashed line between rectangles, denotes one operation from the corresponding operation sets $\mathcal{O} \in \{\mathcal{O}_{a}, \mathcal{O}_{p}, \mathcal{O}_{r},\mathcal{O}_{m} \}$. 
The detailed OPs are given in the following.

\begin{table}[ht]
	\caption{The operations used in our search space.}
%	\vspace{-10px}
	\begin{tabular}{l|L{170pt}}
		\hline
		Module name           & Operations \\ \hline
		Aggregation $\mathcal{O}_a$ & \texttt{GCN}, \texttt{GAT}, \texttt{SAGE}, \texttt{GIN}, \texttt{GRAPHCONV}, \texttt{MLP}           \\ \hline
		Pooling $\mathcal{O}_p$&  \texttt{TOPKPOOL}, \texttt{SAGPOOL}, \texttt{ASAP}, \texttt{HOPPOOL\_1}, \texttt{HOPPOOL\_2}, \texttt{HOPPOOL\_3}, \texttt{MLPPOOL}, \texttt{GCPOOL}, \texttt{GAPPOOL}, \texttt{NONE}            \\ \hline
		Readout $\mathcal{O}_r$& \texttt{GLOBAL\_SORT}, \texttt{GLOBAL\_ATT}, \texttt{SET2SET}, \texttt{GLOBAL\_MEAN}, \texttt{GLOBAL\_MAX}, \texttt{GLOBAL\_SUM},  
		\texttt{ZERO}            \\ \hline
		Merge $\mathcal{O}_m$& \texttt{M\_LSTM}, \texttt{M\_CONCAT}, \texttt{M\_MAX}, \texttt{M\_MEAN}, \texttt{M\_SUM}            \\ \hline
	\end{tabular}

\label{tb-search-space}
\end{table}

\noindent\textbf{Aggregation Module.}
We add five widely used GNNs: GCN~\cite{kipf2016semi}, GAT~\cite{velivckovic2017graph}, GraphSAGE~\cite{hamilton2017inductive} with mean aggregator, GIN~\cite{xu2018powerful} and GraphConv~\cite{morris2019weisfeiler}, which denoted as \texttt{GCN}, \texttt{GAT}, \texttt{SAGE}, \texttt{GIN} and \texttt{GRAPHCONV}. Besides, we incorporate the operation \texttt{MLP}, which applies a two-layer MLP (Multilayer Perceptrons) to update node embeddings without using the graph structure.

\noindent\textbf{Pooling Module.}
The pooling operations in our search space can be unified by a computation process as
\begin{align}
	\label{eq-s}
	\bS&=f_s(\bA, \bH), idx = \texttt{TOP}_k(\bS),\\
	\label{eq-sample}
	\bA'&=\bA(idx,idx), \bH'=\bH(idx,:).
\end{align}
We firstly calculate a node score matrix $\bS \in \mathbb{R}^{N \times 1}$ with a score function $f_s$, 
which is used to evaluate the importance of nodes with different metrics, 
then generate the coarse graph by selecting top-$k$ nodes $idx$ with the function $\texttt{TOP}_k$,
and formulating the coarse graph according to Eq. \eqref{eq-sample}. 

Three existing pooling operations \texttt{TOPKPOOL}~\cite{gao2019graph}, 
\texttt{SAGPOOL}~\cite{lee2019self} and \texttt{ASAP}~\cite{ranjan2020asap} are incorporated in our search space. 
We further provide 6 score functions: \texttt{HOPPOOL\_t} formulates the node scores based on the summation of different powers of the adjacency matrix, which is denoted as
% $\bS=\sum_{i=1}^t\bA^i, t \in \{1,2,3\}$, this can be found in~\cite{ma2020path};
$\bS_u=\sum_{i=1}^t\sum_{v \in \widetilde{\mathcal{N}}(u)}\bA^i_{uv}$;  
\texttt{MLPPOOL}, which denoted as $\bS_u = \sigma(\bW_1\sigma(\bW_0\bh_u))$, uses a 2-layer MLP as the score function; \texttt{GCPOOL} uses \texttt{GRAPHCONV} to generate the node scores which is similar to SAGPool~\cite{lee2019self}; 
%and \texttt{GAPPOOL}, which use $\bS_u=\frac{1}{2}\bm{W} \sum_{v \in \widetilde{\mathcal{N}}(u)}(\bh_u-\bh_v)^2$ as the score function.
and the score function of \texttt{GAPPOOL} can be represented as $\bS_u=\frac{1}{2}\bm{W} \sum_{v \in \widetilde{\mathcal{N}}(u)}(\bh_u-\bh_v)^2$.

Apart from these pooling operations, we also add \texttt{NONE} operation, which means no pooling operation in this layer. 
In this way, we can search for hierarchical and global methods adaptively with PAS, which is more flexible than existing methods.
%Note that we do not include spectral-based methods, like EigenPool~\cite{ma2019graph} or MinCutPool~\cite{bianchi2020spectral}, into the search space, since they are inherently computational expensive and can not be integrated with the used message passing based architecture backbone. 
%The DiffPool~\cite{ying2018hierarchical} and GMN~\cite{khasahmadi2020memory} methods group nodes into new clusters, and these clusters are treated as new nodes in the coarsen graph. Therefore, the graphs before and after pooling can be merge into one graphs by constructing the edges among the nodes of two graphs according to the assignment functions. Thus, these methods can be incorporated with Eq.~\eqref{eq-sample} and they can be treated as a kind of data-augmentation variants of this Pooling Module.
%\huan{what do you mean by data augmentation variant of pooling module? How about we say that for the sake of computational resource, we do not include DiffPool, GMN, EigenPool, and MinCutPool, despite that they are all hierarchical methods?} 
%Considering that these methods are usually computational expensive, we do not incorporate these methods and will it for future work. So is the reason for not choosing the spectral-based methods, like EigenPool~\cite{ma2019graph} or MinCutPool~\cite{bianchi2020spectral}.
%due to the processing of Laplacian matrix

\noindent\textbf{Readout Module.}
We provide 7 global pooling functions to obtain the graph representation vector $\bz \in \mathbb{R}^d$: 3 existing methods \texttt{GLOBAL\_SORT}~\cite{zhang2018end}, \texttt{GLOBAL\_ATT}~\cite{li2015gated} and \texttt{SET2SET}~\cite{vinyals2015order}; simple global mean, max and sum functions denoted as \texttt{GLOBAL\_MEAN}, \texttt{GLOBAL\_MAX} and \texttt{GLOBAL\_SUM} respectively;   
\texttt{ZERO} operation, which generate a zero vector, indicating the graph embeddings in this layer are not used for the final representation.
 
\noindent\textbf{Merge Module.}
Motivated by~\cite{xu2018representation,chen2019powerful} that intermediate layers help to fomulate expressive embeddings, we add 5 merge functions to incorporate the graph representations in each layer: LSTM, concatenation, max, mean and sum,
%intermediate layers helps to fomulate expressive embeddings. Thus, we add readout module in each layer and use 
%LSTM, concatenation, max, mean and sum operations 
%to merge the graph representations of all intermediate layers, e.g., $\bz_0$ to $\bz_2$ in Figure~\ref{fig-framework}(a). 
which denoted as \texttt{M\_LSTM}, \texttt{M\_CONCAT}, \texttt{M\_MAX}, \texttt{M\_MEAN} and \texttt{M\_SUM} in our search space. 

%\vspace{1pt}
%\noindent\textbf{Discussions.}
{
\color{blue}
%Compared with existing methods in Table \ref{tb-compare-existing}, PAS provides a novel search space consisting of 4 modules, which can cover most existing methods in this search space. 
%More operations can be trivially added into this search space for further graph classification architectures learning, 
%\sout{e.g., the aggregation operations in GraphNAS and AutoGM. }
%In this sense, the capability of the PAS framework can be further improved.

%With the design of these 4 modules, PAS enables the data-specific pooling architectures learning for graph classification task. Especially, 
%PAS can generate different combinations and numbers of pooling operations, on the contrary, existing methods only design architectures on fixed pooling schema, e.g., use fixed pooling operation in each layer~\cite{ranjan2020asap} or only global pooling function directly~\cite{zhang2018end}.
}

As shown in Figure~\ref{fig-framework}(b), the example of a 2-layer architecture backbone, 
%there are 2 Aggregation Modules, 2 Pooling Modules, 3 Readout Modules, and 1 Merge Module, 
the search space size is $6^2 \times 10^2 \times 7^3 \times 5 = 6,174,000$. 
With so many candidate architectures in the search space, PAS can generate data-specific pooling architectures beyond existing human-designed ones (see Figure \ref{fig-appendix-searched-archs}). 
Moreover, more operations can be trivially added to the search space to enlarge the search space if the computational budget is enough, e.g., DiffPool~\cite{ying2018hierarchical}, GMN~\cite{khasahmadi2020memory}, STRUCTPOOL~\cite{yuan2019structpool}, EigenPool~\cite{ma2019graph}, etc. Therefore, it also means that an efficient search method is needed over such a large search space.

%different combinations and numbers of pooling operations which are not explored with human experience. On the contrary, existing methods only design architectures on fixed pooling schema, e.g., use same pooling operation in each layer~\cite{ranjan2020asap} or only global pooling function directly~\cite{zhang2018end}.
%Based on the unified framework, more operations can be trivially added into this search space for further graph classification architectures learning. In this sense, the capability of the PAS framework can be further improved. 

\subsection{Differentiable Search Algorithm}
%\subsection{Pooling Module Relaxation}
%\wei{3 subsections: stochastic relaxation + mixed pooling operations + pipeline of search stage.}

%%\footnote{+qm+ should this Section~be split into two
%	1) design of the supernet
%	and 2) search algorithm.
%	note that you only use 1 paragraph to talk about the search .\huancheck{}}
%%\footnote{+qm+ what's new here?
%	at least what is new based on NAS for GNN (Section~\ref{sec-nas-gnn})? \huan{mixed pooling OP here is new in NAS for GNN.}}
%%Following the existing over-parameterized network 
%%\footnote{+qm+ you have not talked about these works in related works,
%	so, instead of say ``following xxx'',
%	it is better finish description of your own work and then
%	compared with these existing ones later.\huancheck{}}

%\cite{xie2018snas,cai2018proxylessnas,yao2019differentiable,pham2018efficient},
%\subsubsection{Continuous relaxationof the search space}
%\noindent\textbf{Continuous relaxation in differentiable search algorithm.}
In this part, we develop a differentiable search algorithm. 
Specially, the designed coarsening strategy is detailed, 
which makes it feasible to continuously relax the selection of pooling operations.
%, thus the search space.

%The designed search space is too large, so that it is inefficiency to sample architectures with human experience or RL and EA based methods~\cite{gao2019graphnas,jiang2020graph}. 
%
Technically speaking, as done in existing NAS works \cite{liu2018darts,xie2018snas,zhao2021search}, one needs to relax the the search space into continuous one, thus the discrete selection of operations is relaxed by a weighted summation of all possible operations as
\begin{equation}
	\label{eq-supernet-weightedsum}
	\bar{o}(x)
	= \sum\nolimits_{i=1}^{\left|\mathcal{O}\right|} c_io_i(x),
	\vspace{-1pt}
\end{equation}
where $x$ denotes the input representation of each module, and $c_i \in (0,1)$ denotes the weight of the $i$-th operation $o_i(\cdot)$ in the set $\mathcal{O}$.
It is generated by one reparameterization trick, which is used as continuous relaxation function in NAS, as $c_i=g(\bm{\alpha}), \bm{\alpha} \in \mathbb{R}^{\left|\mathcal{O}\right|}$ and $\alpha_i$ is the corresponding learnable parameter for $c_i$.

However, it is non-trivial to apply existing differentiable search algorithms in pooling architecture search.
Different pooling operations produce diverse coarse graphs consisting of different nodes and edges as shown in Figure~\ref{fig-framework}(c).
How to relax the discrete coarsen graphs into continuous is still not solved in NAS. 
Here, to utilize the differentiable search algorithm, we design one coarsening strategy to address the pooling module relaxation problem.

%\paragraph{Coarsening strategy.}
\noindent\textbf{Coarsening strategy.}
For Aggregation, Readout and Merge Module in the supernet, 
we can generate the mixed results with relaxation function $g(\cdot)$ by constraining the node embedding dimension as shown in 
\begin{align}
\label{eq-agg}
\bH^{la}&=\sum\nolimits_{i=1}^{\left|\mathcal{O}_a\right|} c_i^{al}o_i(\bA^{l-1},\bH^{l-1}),\\
\label{eq-readout}
\bz^l&=\sum\nolimits_{i=1}^{\left|\mathcal{O}_r\right|} c_i^{rl}o_i(\bA^l, \bH^l),\\
\label{eq-merge}
\textbf{z}^F &=\sum\nolimits_{i=1}^{\left|\mathcal{O}_m\right|}c_i^{m} o_i(\textbf{z}^0,\textbf{z}^1,\cdots,\textbf{z}^L),
\end{align}
where $c_i^{al}$ and $c_i^{rl}$ denote the weights of $i$-th aggregation operation and global pooling operation in $l$-th layer, $c_i^m$ denotes the $i$-th operation in the Merge Module.

\begin{algorithm}[t]
	\caption{PAS - Pooling Architecture Search}
	%	\wei{add ref to sec 3.2.}
	\begin{algorithmic}[1]
		\Require{Training dataset $\mathcal{D}_{train}$, validation dataset $\mathcal{D}_{val}$, the epoch $T$ for search}
		\Ensure{The searched architecture.}
		\State Random initialize the parameters $\bm{\alpha}$ and $\bW$.
		%		\State Sample a set of minibatchs $\mathcal{G}=\{\mathcal{G}_1,\ldots,\mathcal{G}_b,\ldots\}$ from $\mathcal{D}_{train}$.\wei{hebing line 2, 4.} 
		\While {$t=1,\ldots,T$}
		\For {each minibatch $\mathcal{G}_b \in \mathcal{D}_{train}$}
		\State Calculate the operation weightes $\textbf{C} = g(\bm{\alpha})$.
		%		\For {each graph $(G=(\bA^0,\bH^0),y) \in \mathcal{G}_b$}
		%		\State $z^0=\sum_{i=1}^{\left|\mathcal{O}_r\right|} c_i^{r0}o_i(\bA^0, \bH^0)$ 
		%		\State Calculate $\bz^0$ with Eq.~\eqref{eq-readout}
		%		\For {$l=1$ to $L$}
		%		\State Calculate $\bH^{la}, G^l$ with Eq.~\eqref{eq-agg} and \eqref{eq-weightsum-pool}.
		%		\State Calculate $\bz^l$ with Eq.~\eqref{eq-readout}.
		%		\State $\bH^{la}=\sum_{i=1}^{\left|\mathcal{O}_a\right|} c_i^{al}o_i(\bA^{l-1},\bH^{l-1})$, $//$Aggregation \label{alg-agg}
		%		\State  $\textbf{A}^{l} = \sum_{i=1}^{\left|\mathcal{O}_p\right|}c_i^{pl}o_i(\bA^{l-1},\bH^{la})$, \label{alg-pooling}
		%		\Statex  $\qquad\qquad\qquad\textbf{H}^{l} = \sum_{i=1}^{\left|\mathcal{O}_p\right|}c_i^{pl}o_i(\bA^{l-1},\bH^{la})$,  $//$ Pooling
		%
		%		\State $\bz^l=\sum_{i=1}^{\left|\mathcal{O}_r\right|} c_i^{rl}o_i(\bA^l, \bH^l)$,  $//$ Readout \label{alg-readout}
		%		\EndFor
		%		\State $\textbf{z}^F =\sum_{i=1}^{\left|\mathcal{O}_m\right|}c_i^{m} o_i(\textbf{z}^0,\textbf{z}^1,\cdots,\textbf{z}^L)$,  $//$ Merge \label{alg-merge}
		%		\State Calculate $\bz^F$ with Eq.~\eqref{eq-merge}.
		%		\State $p=\textbf{W}_{cl}^T\textbf{z}^F + b$.
		%		\EndFor
		%		\State $\mathcal{L}_{train}=-\frac{1}{M_t}\sum_{m=1}^{M_t}\left[y_m\text{log}(p_m)+(1-y_m)\text{log}(1-p_m)\right]$ , //cross-validation loss
		%		\State $\mathcal{L}=\frac{1}{\left|\mathcal{G}_b\right|}\sum_G LOSS(p, y)$
		\State Calculate the graph representation $\bz^F$ for each graph.
		\State Update $\bW$ with training loss $\mathcal{L}_{train}$.
		\EndFor
		\For {each minibatch $\mathcal{G}_b \in \mathcal{D}_{val}$}
		\State Calculate the operation weightes $\textbf{C} = g(\bm{\alpha})$.
		\State Calculate the graph representation $\bz^F$ for each graph.
		\State Update $\bm{\alpha}$ with validation loss $\mathcal{L}_{val}$.
		\EndFor
		\EndWhile
		\State Preserve the operation with the largest weight in each module.\\
		\Return The searched architecture.
	\end{algorithmic}
	\label{alg-pas}
\end{algorithm}

However, it is infeasible to directly compute the weighted summation of the output of all operations with Eq.~\eqref{eq-supernet-weightedsum} in the Pooling Module since 
the coarse graphs generated by different pooling operations contain diverse nodes and edges.
As shown in Figure~\ref{fig-framework}(c), the different node sets $\{0,1,4,5\}$ and $\{0,1,2,5\}$ are preserved by two pooling operations, and the features matrix $\bH^l_i \in \mathbb{R}^{4\times d}$ can not directly be added as the aggregation module in Eq.~\eqref{eq-agg} due to the node mismatch problem, the edges are the same. 
Thus, it is unachievable to relax the Pooling Module, and the usage of gradient descent on this supernet is infeasible.

To address the challenge facing the Pooling Module in the supernet computation, 
we design a \textit{coarsening strategy} to make the pooling module relaxation feasible,
thus the supernet can be trained with gradient descent. In this coarsening strategy, we firstly calculate the node score matrix $\bS$ and select the top-$k$ nodes \texttt{idx} with Eq. \eqref{eq-s} for each pooling operation. Rather than formulate the coarse graphs with Eq. \eqref{eq-sample} as in general pooling operation, as shown in Figure~\ref{fig-framework}(c), we generate the result  $G_i^l=(\bA^l_i,\bH^l_i)$ of $i$-th pooling operation by making the coarse graph keep the ``shape'', i.e., the same node numbers as the input graph $G^{la}$. By setting the unselected ``part'' to $0$, i.e., the features of unselected nodes and weights of unselected edges,
different pooling results $G^l_i$ can be added to generate the mixed results of the Pooling Module, which is shown in the following:
\begin{align}
\label{eq-weightsum-pool}
G^l = (\bm{A}^l,\bm{H}^l) 
= (\sum\nolimits_{i=1}^{\left|\mathcal{O}_p\right|}c_i^{pl}\bm{A}_i^l, 
\sum\nolimits_{i=1}^{\left|\mathcal{O}_p\right|}c_i^{pl}\bm{H}_i^{l}),
\end{align}
where $c_i^{pl}$ denotes the weight of $i$-th pooling operation in $l$-th layer.

With the designed coarsening strategy, we can relax the pooling module into continuous 
and allow the usage of gradient descent.
Therefore, it is easy to generate the graph representation $\bz^F$ with Eq.\eqref{eq-supernet-weightedsum} as in Figure~\ref{fig-framework}(b).

%\paragraph{Optimization and deriving process.}
\noindent\textbf{Optimization and deriving process.}
In this paper, we choose the Gumbel-Softmax~\cite{jang2016categorical} as the continuous relaxation, which is designed to approximate discrete distribution in a differentiable manner and shown useful for supernet training in NAS~\cite{xie2018snas,dong2019searching}.
Since the training loss $\mathcal{L}_{train}$, validation loss $\mathcal{L}_{val}$ and the relaxation function are differentiable, we can optimize the parameters $\bm{\alpha}$ and $\bW$ with gradient descent as shown in Alg.~\ref{alg-pas}. 
After finishing the search process, 
we preserve the operations with the largest weights in each module, from which we obtain the searched architecture.

%\paragraph{Discussions.}
%As shown in Table~\ref{tb-compare-existing}, existing NAS methods for GNN use RL, EA and Bayesian as their search algorithms, 
%which require thousands of architecture evaluations and are thus time-consuming to search one data-specific architecture (see Figure~\ref{fig-gpuhours}). 
%In this paper, we propose one coarsening strategy to relax the pooling module into continuous, 
%which provides one platform to applying differentiable search algorithms in pooling architecture search. 
%With the differentiable search algorithm introduced before, PAS is efficient in orders than existing RL-based NAS methods.

%\begin{algorithm}[t]
%	\caption{Framework Results.}
%	\begin{algorithmic}[1]
%		\Require{Input graph $G=(\bA^0,\bH^0)$}
%		\Ensure{The graph representation $\bz^F$.}
%		\State Calculate $\bz^0$ with Eq.~\eqref{eq-readout}
%		\For {$l=1$ to $L$}
%		\State Calculate $\bH^{la}, G^l$ with Eq.~\eqref{eq-agg} and \eqref{eq-weightsum-pool}.
%		\State Calculate $\bz^l$ with Eq.~\eqref{eq-readout}.
%		\EndFor
%		\State Calculate $\bz^F$ with Eq.~\eqref{eq-merge}.\\
%%		\State $p=\textbf{W}_{cl}^T\textbf{z}^F + b$. \label{alg-classifier} \\
%		\Return The graph representation $\bz^F$.
%	\end{algorithmic}
%	\label{alg-pas-single}
%\end{algorithm}

\section{Experiments}
\label{sec-exp}
%In this part, we present the experiments on extensive datasets.
%and then we proof the importance of pooling modules and readout modules. We provide extensive performance comparisons among x datasets to further evaluate the search space in this paper.
\subsection{Experimental Settings}
%\subsubsection{Datasets and baselines}

%\noindent\textbf{Datasets.} We choose 5 datasets, 2 bioinformatics datasets: D\&D and PROTEINS, 1 small molecules datasets: COX2, and 2 social network datasets: IMDB-BINARY and IMDB-MULTI. The statistics of these datasets are given in Table~\ref{tb-graph-dataset}, and more details can be found in Appendix~\ref{sec-appendix-exp}. \huan{Refs for datasets.}
%\noindent\textbf{} 

\noindent\textbf{Datasets.}
In this paper, we use six datasets as shown in Table~\ref{tb-graph-dataset}.
D\&D and PROTEINS datasets, provided by ~\cite{dobsondistinguishing}, are both protein graphs. 
In the D\&D dataset, nodes represent the amino acids and two nodes are connected if the distance is less than 6 $\dot{A}$. In the PROTEINS dataset, nodes are secondary structure elements and edges represent nodes are in an amino acid or in a close 3D space. 
IMDB-BINARY and IMDB-MULTI datasets, provided by~\cite{yanardag2015deep}, are movie-collaboration datasets that contain the actor/actress and genre information from IMDB. Nodes represent the actor/actress and edges mean they appear in the same movies.
COX2 dataset, provided by~\cite{sutherland2003spline}, is a set of 467 cyclooxygenase-2 (COX-2) inhibitors and classify these compounds as active or inactive within vitro activities against human recombinant enzyme values.
NCI109 dataset, provided in ~\cite{wale2006comparison,shervashidze2011weisfeiler}, represents two balanced subsets of datasets of chemical compounds screened for activity against non-small cell lung cancer and ovarian cancer cell lines respectively.

\begin{table}[ht]
	%	\small
	\setlength\tabcolsep{1pt}
	\centering
	
	\caption{Statistics of the datasets from three domains. PRO, IMDB-B and IMDB-M are short for PROTEIN, IMDB-BINARY and IMDB-MULTI, respectively.}
	\label{tb-graph-dataset}
	\vspace{-10pt}
	\begin{tabular}{c|C{25pt}|C{28pt}|C{28pt}|C{30pt}|C{30pt}|c}
		\toprule
		Dataset     & \# of Graphs & \# of Feature & \# of Classes & Avg.\# of Nodes & Avg.\#  of Edges & Domain     \\ \midrule
		D\&D        & 1,178          & 89           & 2       & 384.3              & 715.7              & Bioinfo\\ \hline
		PRO    & 1,113          & 3            & 2       & 39.1               & 72.8               & Bioinfo \\ \hline
		IMDB-B & 1,000          & 0            & 2       & 19.8               & 96.5               & Social         \\ \hline
		IMDB-M  & 1,500          & 0            & 3       & 13                 & 65.9               & Social         \\ \hline		
		COX2        & 467            & 3            & 2       & 41.2               & 43.5               & Chemistry      \\ \hline 
		NCI109  &4127 &0 &2& 29.69 &32.13 &Chemistry \\ 
		\bottomrule
	\end{tabular}
\end{table}

%\noindent\textbf{} 

\noindent\textbf{Baselines.}
We use 3 types of baselines: global pooling methods, hierarchical pooling methods and NAS methods for GNNs. 

%For global pooling methods, we add a readout function at the end of GCN, GAT, GraphSAGE and GIN methods. We also provide the skip connection version on the former 4 methods which denoted as '-JK'.

%For global pooling methods, we add a readout function at the end of GCN, GAT, GraphSAGE and GIN methods. For further comparison, we also add extra 4 baselines that combine with JK-Net~\cite{xu2018representation} on the former 4 methods. Based on GraphSAGE, we add \texttt{GLOBAL\_SORT} and \texttt{GLOBALl\_ATT} readout functions corresponding to \texttt{DGCNN} and \texttt{Global\_attention} baselines in Table~\ref{tb-performance-comparisons}. Apart from the above GNN methods, we add the \texttt{MLP\_baseline}, a single-layer MLP operation in each layer and a readout function to generate the graph representations.

For global pooling methods, we add a global pooling function at the end of existing node classification methods, e.g., GCN, GAT, SAGE, GIN, and combining with JK-Network~\cite{xu2018representation} which add skip connections to these models, we can formulate another 4 methods which denoted as GCN-JK, GAT-JK, SAGE-JK, GIN-JK, respectively.
We add the existing global pooling method DGCNN~\cite{zhang2018end} in our experiment and MLP-Baseline which only contains MLPs and one global pooling function. 

For hierarchical pooling methods, we use 5 popular ones: Graph U-Net~\cite{gao2019graph}, DiffPool~\cite{ying2018hierarchical}, SAGPool~\cite{lee2019self}, ASAP~\cite{ranjan2020asap} and MinCutPool~\cite{bianchi2020spectral}.

%We use 5 popular hierarchical pooling methods: Graph U-Net, K-GNN, DiffPool, SAGPool, and ASAP.

\begin{table*}[t]
	\small
	\centering
	\caption{Performance comparisons of PAS and all baselines. We report the mean test accuracy and the standard deviation by 10-fold cross-validation. The best results in different groups of baselines are underlined, and the best result on each dataset is in boldface.}
	\label{tb-performance-comparisons}
	\vspace{-10pt}
%	\begin{tabular}{C{42pt}|c|c|c|c|c|c}
%\multirow{10}{*}{\begin{minipage}{42pt}Global pooling\end{minipage}} 
\begin{tabular}{C{42pt}|c|c|c|c|c|c|c}
	\cline{1-8}
	& Method       & D\&D                                                                  & PROTEINS                                                               & IMDB-BINARY                                    & IMDB-MULTI                                                             & COX2                                           & NCI109            \\ \cline{1-8}
	\multirow{10}{*}{\begin{minipage}{42pt}Global\\pooling\end{minipage}}      
	& GCN          & \underline{0.7812$\pm$0.0433}                                                     & 0.7484$\pm$0.0282                                                      & 0.7267$\pm$0.0642                              & 0.5040$\pm$0.0302                                                      & 0.7923$\pm$0.0219                              & 0.7344 $\pm$   0.0192 \\ 
	& GAT          & 0.7556$\pm$0.0372                                                     & 0.7530$\pm$0.0372                                                      &  \underline{0.7407$\pm$0.0453} & 0.4967$\pm$0.0430                                                      & 0.8156$\pm$0.0417                              & 0.7410 $\pm$   0.0245 \\ 
	& SAGE         & 0.7727$\pm$0.0406                                                     & 0.7375$\pm$0.0297                                                      & 0.7217$\pm$0.0529                              & 0.4853$\pm$0.0543                                                      & 0.8031$\pm$0.0594                              & \underline{0.7553 $\pm$ 0.0164}   \\ 
	& GIN          & 0.7540$\pm$0.0368                                                     & 0.7448$\pm$0.0278                                                      & 0.7167$\pm$0.0277                              & 0.4980$\pm$0.0250                                                      &  \underline{0.8309$\pm$0.0417} & 0.7456 $\pm$   0.0210 \\ \cline{2-8}
	& GCN-JK       & 0.7769$\pm$0.0235                                                     &  \underline{0.7539$\pm$0.0517}                         & 0.7347$\pm$0.0445                              & 0.4886$\pm$0.0496                                                      & 0.7966$\pm$0.0226                              & 0.7383 $\pm$ 0.0188   \\ 
	& GAT-JK       &  0.7809$\pm$0.0618                        & 0.7457$\pm$0.0405                                                      & 0.7327$\pm$0.0468                              & 0.4953$\pm$0.0384                                                      & 0.8072$\pm$0.0287                              & 0.7172 $\pm$ 0.0292   \\ 
	& SAGE-JK      & 0.7735$\pm$0.0420                                                     & 0.7494$\pm$0.0355                                                      & 0.7287$\pm$0.0618                              & 0.4973$\pm$0.0340                                                      & 0.7990$\pm$0.0415                              & 0.7325 $\pm$ 0.0356   \\ 
	& GIN-JK       & 0.7513$\pm$0.0395                                                     & 0.7312$\pm$0.0451                                                      & 0.7207$\pm$0.0486                              &  \underline{0.5080$\pm$0.0302}                         & 0.8117$\pm$0.0467                              & 0.7441 $\pm$0.0220    \\ 
	& DGCNN        & 0.7666$\pm$0.0403                                                     & 0.7357$\pm$0.0469                                                      & 0.7367$\pm$0.0570                              & 0.4900$\pm$0.0356                                                      & 0.7985$\pm$0.0264                              & 0.7506 $\pm$ 0.0165   \\ 
	& MLP-Baseline & 0.7752$\pm$0.0390                                                     & 0.7239$\pm$0.0353                                                      & 0.7287$\pm$0.0520                              & 0.4980$\pm$0.0428                                                      & 0.7838$\pm$0.0104                              & 0.6445 $\pm$ 0.0160   \\ \cline{1-8}
	& ASAP         & 0.7735$\pm$0.0415                                                     & 0.7493$\pm$0.0357                                                      & 0.7427$\pm$0.0397                              &  \underline{0.5013$\pm$0.0344}                         &  \underline{0.8095$\pm$0.0320} & 0.7376 $\pm$ 0.0224   \\ 
	\multirow{3}{*}{\begin{minipage}{42pt}Hierarchical pooling\end{minipage}} & SAGPool      & 0.7506$\pm$0.0506                                                     & 0.7312$\pm$0.0447                                                      &  \underline{0.7487$\pm$0.0409} & 0.4933$\pm$0.0490                                                      & 0.7945$\pm$0.0298                              & 0.6489 $\pm$ 0.0315   \\ 
	& Graph U-Net  & 0.7710$\pm$0.0517                                                     & 0.7440$\pm$0.0349                                                      & 0.7317$\pm$0.0484                              & 0.4880$\pm$0.0319                                                      & 0.8030$\pm$0.0421                              & 0.7279 $\pm$ 0.0229   \\ 
	& DiffPool     & 0.7775$\pm$0.0400                                                     & 0.7355$\pm$0.0322                                                      & 0.7186$\pm$0.0563                              & 0.4953$\pm$0.0398                                                      & 0.7966$\pm$0.0264                              & 0.7315 $\pm$ 0.0214   \\
	& MinCutPool   &  \underline{0.7803$\pm$0.0363}                        &  \underline{0.7575$\pm$0.0380}                         & 0.7077$\pm$0.0489                              & 0.4900$\pm$0.0283                                                      & 0.8007$\pm$0.0385                              & \underline{0.7405$\pm$0.0248}                  \\ \cline{1-8}
	\multirow{7}{*}{NAS}                  & GraphNAS     & 0.7198$\pm$0.0454                                                     & 0.7251$\pm$0.0336                                                      & 0.7110$\pm$0.0230                              & 0.4693$\pm$0.0364                                                      & 0.7773$\pm$0.0140                              & 0.7228 $\pm$   0.0228 \\
	& GraphNAS-WS  & 0.7674$\pm$0.0455                                                     &  \underline{0.7520$\pm$0.0251}                         & 0.7360$\pm$0.0463                              & 0.4827$\pm$0.0350                                                      & 0.7816$\pm$0.0058                              & 0.7049 $\pm$   0.0192 \\ 
	& SNAG         & 0.7223$\pm$0.0386                                                     & 0.7053$\pm$0.0311                                                      & 0.7250$\pm$0.0461                              & 0.4933$\pm$0.0289                                                      & 0.7903$\pm$0.0212                              & 0.7090$\pm$0.0224
	                  \\ 
	& SNAG-WS      & 0.7351$\pm$0.0303                                                     & 0.7233$\pm$0.0244                                                      & 0.7360$\pm$0.0516                              &  \underline{0.5000$\pm$0.0248}                         & 0.8054$\pm$0.0381                              & 0.7063$\pm$0.0160
	                  \\ 
	& EA           & 0.7514$\pm$0.0301                                                     & 0.7341$\pm$0.0298                                                      &  \underline{0.7400$\pm$0.0412} & 0.4860$\pm$0.0405                                                      & 0.7945$\pm$0.0159                              & \underline{0.7324$\pm$0.0126}                   \\
	& Random       &  \underline{0.7792$\pm$0.0482}                        & 0.7394$\pm$0.0423                                                      & 0.7210$\pm$0.0554                              & 0.4980$\pm$0.0398                                                      &  \underline{0.8073$\pm$0.0231} & 0.7306$\pm$0.0241
	                   \\ 
	& Bayesian     & 0.7555$\pm$0.0321                                                     & 0.7314$\pm$0.0239                                                      & 0.7270$\pm$0.0335                              & 0.4980$\pm$0.0397                                                      & 0.8029$\pm$0.0172                              & 0.7204$\pm$0.0114
	                  \\ \cline{1-8}
	& PAS          & \textbf{0.7896$\pm$0.0368} &  \textbf{0.7664$\pm$0.0329} &  \textbf{0.7510$\pm$0.0532}    & \textbf{0.5220$\pm$0.0373} &  \textbf{0.8344$\pm$0.0633}    & \textbf{0.7684$\pm$0.0272}                  \\ \cline{1-8}
\end{tabular}
\vspace{-10pt}
\end{table*}

%For NAS baselines, We first use 2 representive
%As shown in Table~\ref{tb-compare-existing}, existing NAS methods are all in global pooling manner with the lack consideration of the pooling operations.
Existing NAS methods focus on design aggregation and global pooling operations and lack the consideration of the pooling operations.
%focus on design aggregation layers in GNNs. With the lack consideration of the pooling operations, the searched architectures are all in global pooling manner. To evaluate the data-specific global pooling architectures, 
% generate the global pooling architectures merely.
% for node classification task. These methods can be applied into graph classification with one global pooling functions, and lead to one global pooling framework.
%To evaluate these methods in graph classification, 
To evaluate these global pooling NAS methods, we choose 4 representive methods to learn architectures based on the diverse search space and \texttt{GLOBAL\_MEAN} operation. 
%Based on the publicly available code, we design 4 baselines based on these existing methods which design aggregation layers based on the \texttt{GLOBAL\_SUM} operation.
%evaluate the  
%we provide 4 RL based baselines to learn data-specific architectures with the \texttt{GLOBAL\_SUM} operation. 
(a) GraphNAS~\cite{gao2019graphnas}\footnote{https://github.com/GraphNAS/GraphNAS}: a RL based method which design different aggregation operations in each layer; (b) GraphNAS-WS: GraphNAS with the weight sharing schema~\cite{pham2018efficient}; (c) SNAG~\cite{zhao2020simplifying}\footnote{https://github.com/AutoML-4Paradigm/SNAG}: a RL based method which design node aggregations, skip connections and layer aggregations in GNNs; (d) SNAG-WS: SNAG with the weight sharing schema.
The evaluations of other global pooling NAS methods with diverse space and differentiable search algorithms is equivalent to the experiments in Section~\ref{sec-ablation-pool}.
%
%Considering that the code of global pooling method RE-MPNN\cite{jiang2020graph} and differentiable methods~\cite{zhao2021search,li2021one,cai2021rethinking} are not publicly available, we will show the evaluations of these global pooling NAS methods in Section~\ref{sec-ablation-pool}.
%
Since existing methods cannot learn the data-specific pooling operations, we further provide 3 baselines for comparisons based on the proposed search space in Section~\ref{sec-search-space}.
(e) EA based method (EA): update populations with mutation and crossover operations;
(f) Random search (Random): samples an architecture from search space randomly; (g) Bayesian optimization (Bayesian)~\cite{bergstra2011algorithms}: incorporates with prior information, use tree-structured Parzen estimator as the measurement to find a better architecture.
% (g) DARTS: a differentiable architecture search method which use the \texttt{SOFTMAX} relaxation instead.\huan{Remove}
%
In this respect, these 7 NAS baselines can cover the widely used search space and search algorithms, thus existing NAS methods can be fully evaluated in out experiment. 

\noindent\textbf{Implementation details.}
%All the global and hierarchical baselines are tuned individually with hyperparameters like embedding size, learning rate, dropout, etc. 
%
For NAS baselines and PAS, we derived the candidate GNNs from the corresponding search space in the search process. 
All the human-designed GNNs and the searched candidates are tuned individually with hyperparameters like embedding size, learning rate, dropout, etc. We perform 10-fold cross-validation to evaluate the model performance based on the searched hyperparameters and report the averaged test accuracy and the standard deviations over 10 folds. 
Following the existing pooling architectures~\cite{ying2018hierarchical,lee2019self}, 2-layer backbone is chosen in this paper for all NAS baselines and PAS, and more layers can be trivially added along with the increasing number of nodes and edges on new graphs. 
For IMDB-MULTI dataset, which has a small node numbers, we choose the 1-layer backbone instead. 
%The influence of layer numbers can be found in Appendix.
The temperature is denoted $0.2$ in Gumbel-Softmax.
More experimental details of PAS and the experiments of layer numbers are given in Appendix.

 \subsection{Performance Comparisons}
 \label{subsec-exp-performance}
%\huan{Consider to give the size of parameters following Table~\ref{tb-performance-comparisons} in this section.\weicheck{}}

%\begin{figure*}[ht]
%	\subfigure[D\&D]{
%		\includegraphics[width=0.30\linewidth]{./fig/DD_params_acc}
%	}
%	\subfigure[PROTEINS]{
%		\includegraphics[width=0.30\linewidth]{./fig/PROTEINS_params_acc}
%	}
%	\subfigure[IMDB-BINARY]{
%		\includegraphics[width=0.30\linewidth]{./fig/IMDB-BINARY_params_acc}
%	}
%	\vspace{-10pt}
%	\caption{(Best viewed in color) The test accuracy  w.r.t. model size. The searched architectures by PAS can achieve SOTA performance with moderate size in terms of the model parameters.}
%	\label{fig-params_acc}
%		\vspace{-15pt}
%\end{figure*}

The results are given in Table~\ref{tb-performance-comparisons}, from which we can see that there is no absolute winner from human-designed models on all datasets. For example, GCN performs best on D\&D while SAGPool performs best on IMDB-BINARY. Considering that these datasets are from three domains, it demonstrates the need for adaptive pooling architectures for graph classification. 
Besides, we can see that PAS consistently outperforms all baselines on all datasets, which demonstrates the effectiveness of PAS on searching for data-specific pooling architectures for graph classification. 

\begin{figure}[h]
	\subfigure[D\&D]{
		\begin{minipage}[t]{0.5\linewidth}
			\centering
			\includegraphics[width=0.9\linewidth]{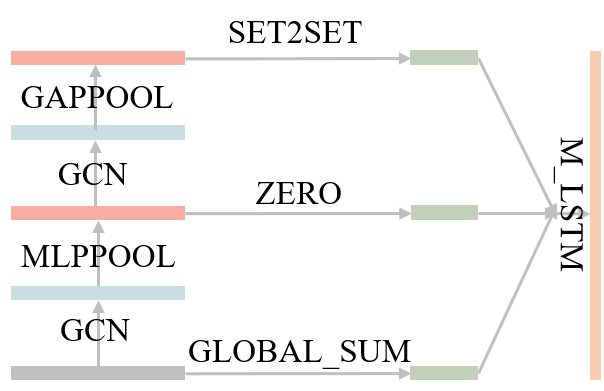}
			%			\caption{D\&D}
		\end{minipage}%
	}%
	\subfigure[PROTEINS]{
		\begin{minipage}[t]{0.5\linewidth}
			\centering
			\includegraphics[width=0.9\linewidth]{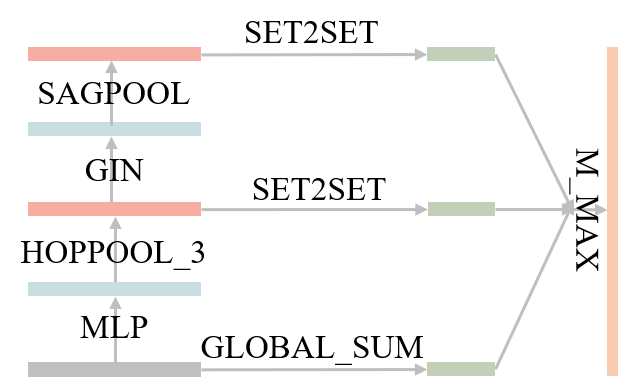}
			%			\caption{D\&D}
		\end{minipage}%
	}\\%
	\subfigure[IMDB-BINARY]{
		\begin{minipage}[t]{0.5\linewidth}
			\centering
			\includegraphics[width=0.9\linewidth]{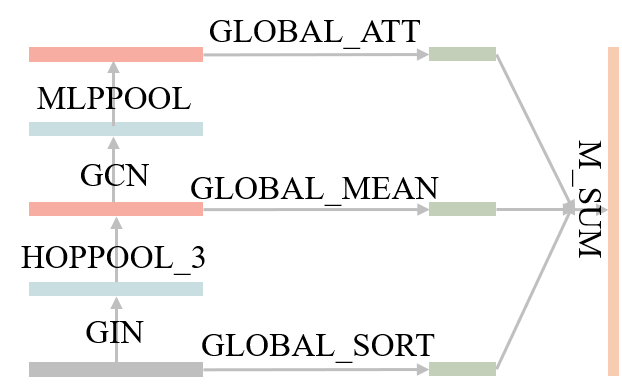}
		\end{minipage}
	}%
	\subfigure[IMDB-MULTI]{
		\begin{minipage}[t]{0.5\linewidth}
			\centering
			\includegraphics[width=0.9\linewidth]{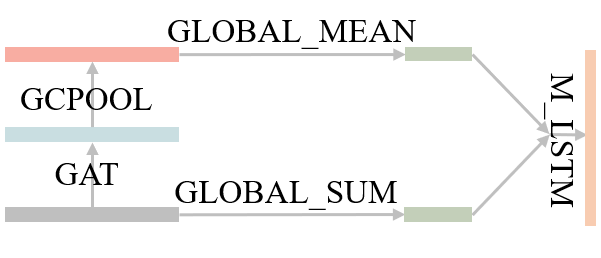}
		\end{minipage}
	}%
	\\
	\subfigure[COX2]{
		\begin{minipage}[t]{0.5\linewidth}
			\centering
			\includegraphics[width=0.9\linewidth]{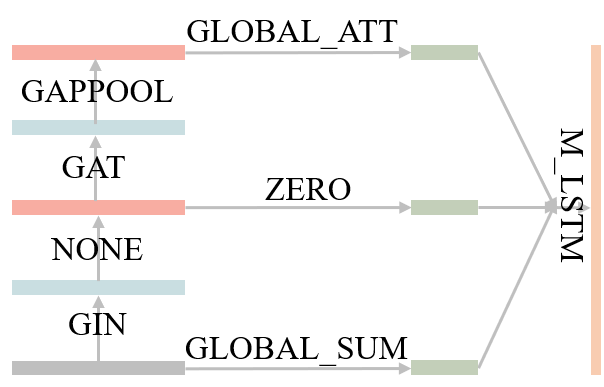}
		\end{minipage}
	}%		
	\subfigure[NCI109]{
	\begin{minipage}[t]{0.5\linewidth}
		\centering
		\includegraphics[width=0.9\linewidth]{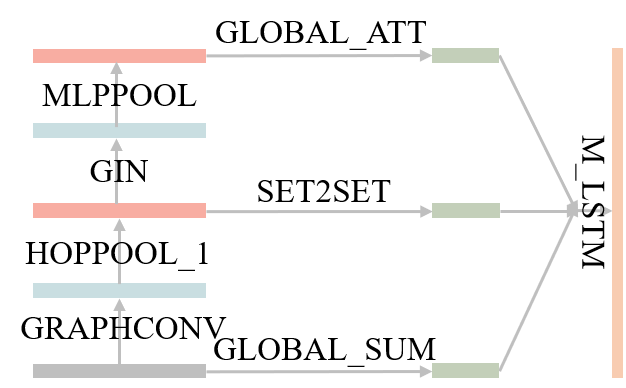}
	\end{minipage}
}%		
%	\vspace{-10pt}
	\caption{The searched architectures on all datasets. We can see that these architectures vary across datasets. Specially, on COX2, the first pooling OP is NONE, which means the model is reduced from a hierarchical pooling method to global pooling method. Based on these results, data-specific pooling architectures can be obtained by PAS.} 
	\label{fig-appendix-searched-archs}
	%\vspace{-15pt}
\end{figure}

When it comes to NAS baselines, the performance gains of PAS are also significant. On one hand, compared to RL-based methods, i.e., GraphNAS and SNAG, the performance gains are mainly from the 4 modules in the designed search space, since GraphNAS and SNAG focus on designing the aggregation layers.
On the other hand, compared with EA, Random and Bayesian, which use the designed search space of PAS, the performance gains are from the differentiable search algorithm on obtaining better architectures. 
		
\begin{figure}[ht]
	\subfigure[D\&D]{
		\includegraphics[width=0.45\linewidth]{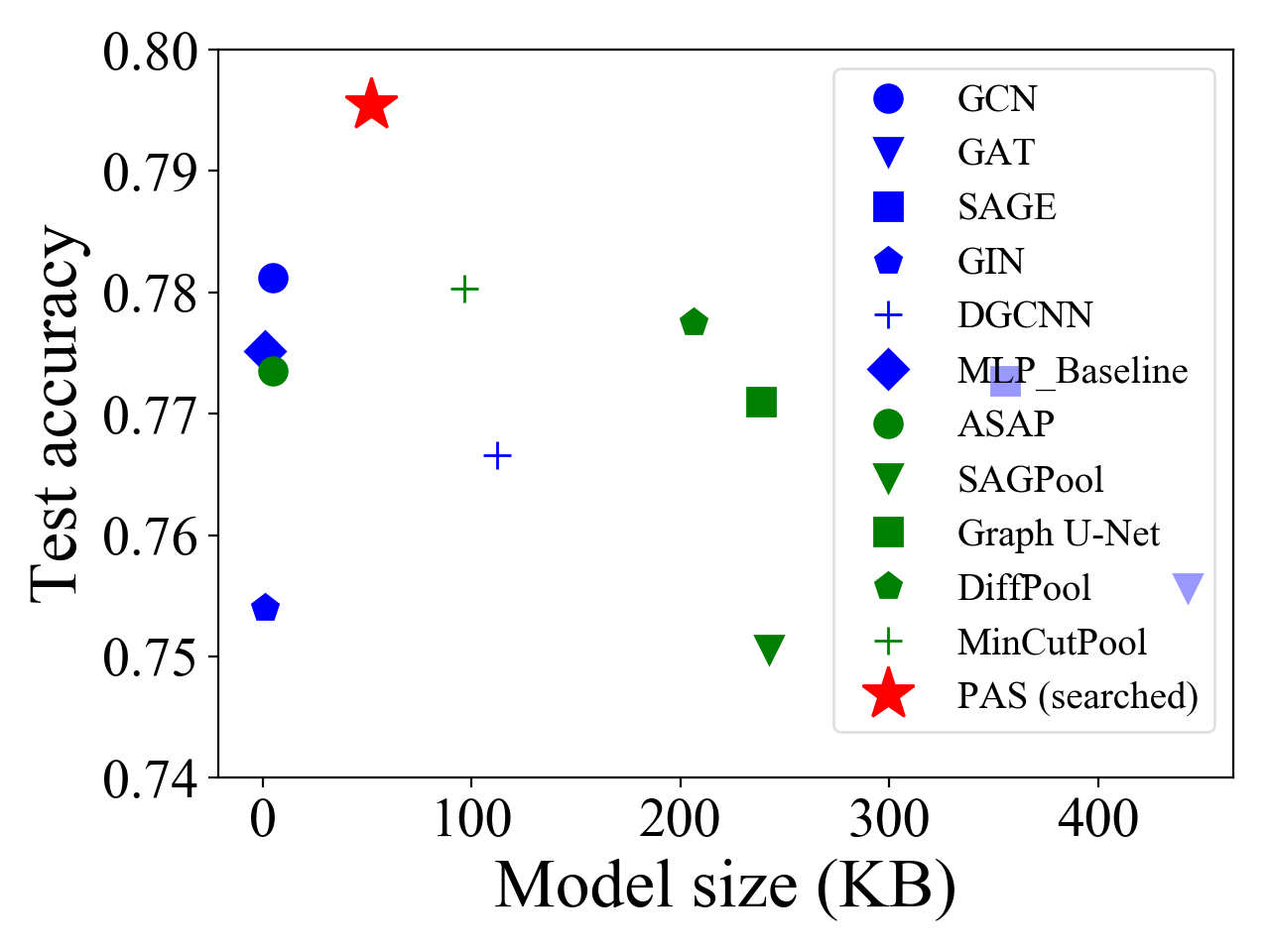}
	}
	\subfigure[PROTEINS]{
		\includegraphics[width=0.45\linewidth]{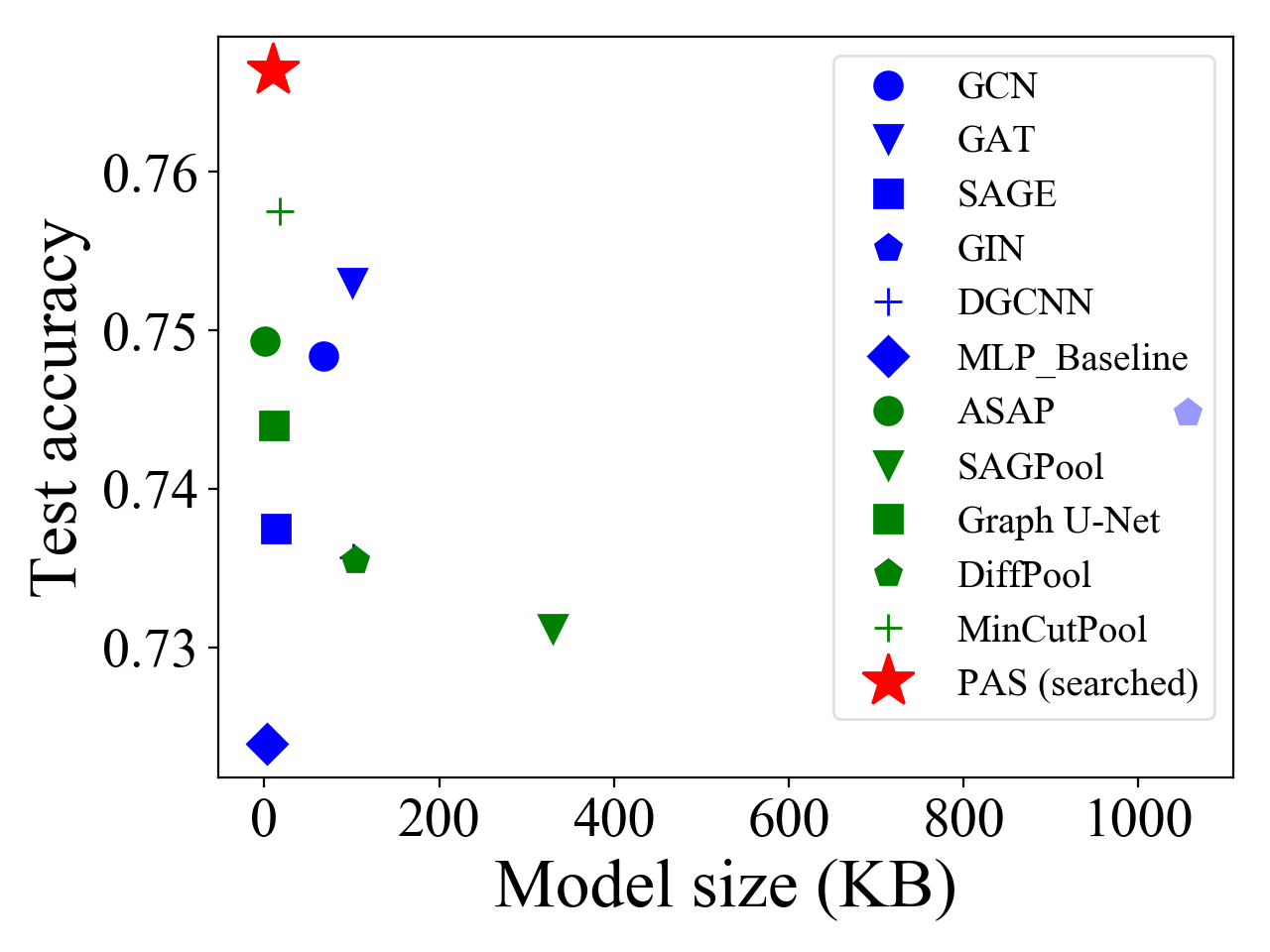}
	}
%	\subfigure[IMDB-BINARY]{
%		\includegraphics[width=0.45\linewidth]{./fig/IMDB-BINARY_params_acc}
%	}
%	\subfigure[IMDB-MULTI]{
%	\includegraphics[width=0.45\linewidth]{./fig/IMDB-MULTI_params_acc}
%	}
%	\subfigure[COX2]{
%	\includegraphics[width=0.45\linewidth]{./fig/COX2_params_acc}
%}
%\subfigure[IMDB-MULTI]{
%	\includegraphics[width=0.45\linewidth]{./fig/IMDB-MULTI_params_acc}
%}
	\vspace{-10pt}
	\caption{(Best viewed in color) The test accuracy  w.r.t. model size. The searched architectures by PAS can achieve SOTA performance with moderate size in terms of the model parameters.}
	\label{fig-params_acc}
	\vspace{-15pt}
\end{figure}

Further, we visualize the searched architectures for all datasets in 
%%\footnote{+qm+ what is HOPPOOL and GCPOOL in the figure?\weicheck{}\wei{introduced in the search space.}}
Figure~\ref{fig-appendix-searched-archs}, from which it is clear that different operation combinations of these four modules are obtained on all datasets, i.e., data-specific architectures. Especially, a global pooling architecture in COX2 is obtained by PAS, since \texttt{NONE} is selected in the Pooling Module of the first layer, while a hierarchical pooling architecture is obtained on D\&D dataset. This observation further indicates the flexibility of the designed search space of PAS.
Besides, we show the test accuracy and model size comparisons among these methods in Figure~\ref{fig-params_acc}. Compared with baselines, the searched architectures, which are shown in Figure~\ref{fig-appendix-searched-archs} and denoted as ``PAS (searched)'' in Figure~\ref{fig-params_acc}, can achieve the SOTA performance with moderate size in terms of the model parameters. It indicates the effectiveness of our method in finding the expressive pooling architectures. More results can be found in Appendix.
%%\footnote{+qm+ still, some artifacts graph can be better for visualization.}

Therefore, these results demonstrate the need for data-specific methods for graph classification, and at the same time, the effectiveness of PAS on designing adaptive pooling architectures.

%\begin{figure}[t]
%	
%	\subfigure[D\&D]{
%		\begin{minipage}[t]{0.5\linewidth}
%			\centering
%			\includegraphics[width=0.9\linewidth]{./fig/searched_dd}
%			%			\caption{D\&D}
%		\end{minipage}%
%	}%
%	\subfigure[COX2]{
%		\begin{minipage}[t]{0.5\linewidth}
%			\centering
%			\includegraphics[width=0.9\linewidth]{./fig/searched_cox2}
%			%			\caption{PROTEINS}
%		\end{minipage}%
%	}%
%	\caption{The adaptive pooling architectures we searched. With the operation \texttt{NONE}, the architectures on COX2 can be reduced into a global pooling method. 
%		%		We search a hierarchical architecture on D\&D dataset, and a global pooling architecture on COX2. 
%}
%	\label{fig-searched_arch}
%	%	\vspace{-10pt}
%\end{figure}
\begin{table}[t]
	\centering
	\caption{Performance of PAS using different search spaces. The first column represents the corresponding module we try to evaluate by fixing it with one OP in the reduced search space.}
	\vspace{-10pt}
	\label{tb-space-evaluation}
	\begin{tabular}{c|l|c|c}
		\toprule
		Fixed & & D\&D                                  & IMDB-MULTI     \\ \hline
		\multirow{2}{*}{Aggregation}	&PAS-GCN     & 0.7835$\pm$0.0407 & 0.5027$\pm$0.0409                     \\ \cline{2-4}
		&PAS-GAT     & 0.7878$\pm$0.0376 & 0.5087$\pm$0.0417                     \\ \hline
		Pooling&	PAS-Global  & 0.7708$\pm$0.0330 & 0.5173$\pm$0.0447                     \\ \hline
		Readout& 	PAS-FR     & 0.7436$\pm$0.0472 & 0.5033$\pm$0.0436                     \\ \hline
		Merge & 	PAS-RM     & 0.7682$\pm$0.0336 & 0.5047$\pm$0.0380                     \\ \hline
		& 	PAS        & \textbf{0.7896$\pm$0.0368} & \textbf{0.5220$\pm$0.0373} \\ \bottomrule
	\end{tabular}
%		\vspace{-15pt}
\end{table}

\subsection{Ablation Studies on the Search Space}
\label{sec-evaluate-space}

%In this section, we first show the pooling architecture evaluation results based on GraphGym benchmark to identify the necessity of pooling architecture. Then we conduct ablation studies to show the influences of the four modules in the proposed method. For simplicity, we use two datasets: D\&D and IMDB-MULTI, and run PAS over different variants of search space, for which the results are shown in Table~\ref{tb-space-evaluation}. 
We conduct ablation studies to show the influences of the four modules in the search space. For simplicity, we use two datasets: D\&D and IMDB-MULTI, and run PAS over different variants of search space, for which the results are shown in Table~\ref{tb-space-evaluation}. 
\subsubsection{Aggregation Module}

To evaluate how the Aggregation Module affects the performance, we only search for the other three modules based on fixed aggregators \texttt{GCN} and \texttt{GAT}, which denoted as PAS-GCN and PAS-GAT, respectively. As shown in Table~\ref{tb-space-evaluation}, with fixed aggregators, PAS-GCN and PAS-GAT have a performance drop compared with PAS. 
Besides, PAS-GAT has a better performance than PAS-GCN, which is consistent with existing works~\cite{velivckovic2017graph}.

This observation demonstrates the importance of including Aggregation Module in the search space, which can also explain the superiority of PAS over human-designed hierarchical pooling methods in Table~\ref{tb-performance-comparisons}, e.g. DiffPool and SAGPool, both of which use fixed aggregation functions. Especially, DiffPool tries to learn the pooling functions. Thus, it shows that the aggregation operations should also be data-specific for graph classification.

%Compare with these methods using fixed aggregators, e.g., DiffPool, SAGPool, PAS-GCN, and PAS-GAT, PAS can learn the interplay among 4 modules, especially the interplay between aggregation and pooling operations. The SOTA performance indicates that learn 4 modules collaboratively is a better pattern for the graph classification task.

\subsubsection{Pooling Module}
\label{sec-ablation-pool}
%Pooling Module is the unique part for graph classification task, it plays an important role in learning graph-level representations, and balabala\textcolor{red}{extra proof of the importance of pooling operations in graph classification task.}. 

%We have shown the Graphgym evaluation results of pooling operations in Figure~\ref{fig-graphgym-winning-ratio}, which indicate that the hierarchical methods have a larger probability to rank first than those global methods due to the pooling operations.
% It owes to the Pooling Module that provides one extra dimension for hierarchical methods to learn expressive graph-level information.  
% \huan{Is this sentence necessary here?}\weicheck{}
%pooling operations provide more opportunities for hierarchical pooling method to have a higher performance than global pooling methods.

Due to the introduction of \texttt{NONE} operation in the Pooling Module in the search space, PAS can automatically emulate both hierarchical and global pooling architectures, which is one of the advantages of the proposed method. As shown in Figure~\ref{fig-appendix-searched-archs}(e), the searched architecture for COX2 correspond to a global pooling method.

%Our search space can integrate both global and hierarchical methods due to the operation \texttt{NONE}, e.g., the searched architectures shown in Figure~\ref{fig-appendix-searched-archs}. PAS can seek for different pooling combinations, rather than use single pooling operations in each layer, or directly use one pooling operations as global pooling baselines.

%\wei{However, in Table~\ref{tb-performance-comparisons}, hierarchical methods have less strength over global pooling methods. It can be explained as: incorporate with more graph-lavel information, Hierarchical methods provide a lower bound for graph classification.}

%\huan{What do you mean by ``only preserving NONE?'' From my understanding, you just keep the global meas/sum in the final layer by removing pooling OPs in other layers.} \wei{Remove pooling operations, the readout module keep same, e.g., 3 readout module in 2-layer backbone.}

Then to evaluate the usefulness of the Pooling Module in the search space, 
%we remove the Pooling Module in the search space 
we search for the combinations of operations in the other three modules based on fixed pooling operation $\texttt{NONE}$. Thus it is equivalent to the global pooling method.
%To evaluate the usefulness of the Pooling Module, we only preserve the \texttt{NONE} operation in the search space, which reduces the architecture backbone to a global pooling by searching for combinations of operations in the other three modules. We 
As shown in Table~\ref{tb-space-evaluation}, the performance of PAS-Global drops significantly compared to PAS, which demonstrates the importance of the designed Pooling Module in the search space. 
In other words, the hierarchical information is useful for obtaining high-quality graph-level representation, which has been demonstrated in existing works, like DiffPool or SAGPool.
More interestingly, speaking of the global pooling manner, PAS-Global can be treated as one representive method of global pooling NAS methods, e.g., 
SANE~\cite{zhao2021search}, DSS~\cite{li2021one} and GNAS~\cite{cai2021rethinking} which design aggregation layers with the differentiable search algorithm based on one global pooling function, and RE-MPNN~\cite{jiang2020graph} which design aggregation layers and global pooling functions with EA.
%RE-MPNN~\cite{jiang2020graph} which use EA as the search algorithm for graph regression and the differentiable methods~\cite{zhao2021search,cai2021rethinking,li2021one} used in graph classification. 
Thus, the performance drop on PAS-Global can demostrates the superiority of PAS over those global pooling NAS methods.

\subsubsection{Readout and Merge Module} 
\label{sec-ablation-readout-merge}
In this section, we evaluate the proposed Readout and Merge Module, which are novel compared to existing pooling architectures. By fixing the global pooling function as \texttt{GLOBAL\_MEAN}, we create the variant PAS-FR, which means that we do not search for different global pooling functions. By removing the Merge Module, we only preserve the last global pooling function, whose output is used as the graph representation $\bz^F$. This variant is denoted by PAS-RM, which means the outputs of intermediate layers are not used. From Table~\ref{tb-space-evaluation}, we can see that
%In PAS, we employ a set of readout functions and merge these graph representation vectors with Merge Module. To evaluate these 2 modules, we set 2 ablation studies. 
\begin{itemize}[leftmargin=*]
	\item The performance drop of PAS-FR means that it is far from satisfying to use a simple mean function to generate fixed-size representation out of all nodes in a graph. As shown in Figure~\ref{fig-appendix-searched-archs}, complex global pooling functions like \texttt{GLOBAL\_ATT} and \texttt{SET2SET} are selected on real-world datasets.
	%	\item 1 fixed readout function (PAS-1F): We remove Merge Module and only use the \texttt{GLOBAL\_MEAN} function to produce the final graph representation $\bz^2$ based on $G^2$;
	\item The performance drop of PAS-RM means that the outputs of intermediate layers are important for the final representation, which have been shown in previous works~\cite{xu2018representation,chen2019powerful}. Thus, it demonstrates the importance of the proposed Merge Module.
\end{itemize}

Taking all results in Table~\ref{tb-space-evaluation} into consideration, we can see that it is important for graph classification to search for combinations of operations from the four essential modules by PAS, which demonstrates the contribution of the designed search space.

\begin{figure}[t]
	\centering
	\includegraphics[width=0.8\linewidth]{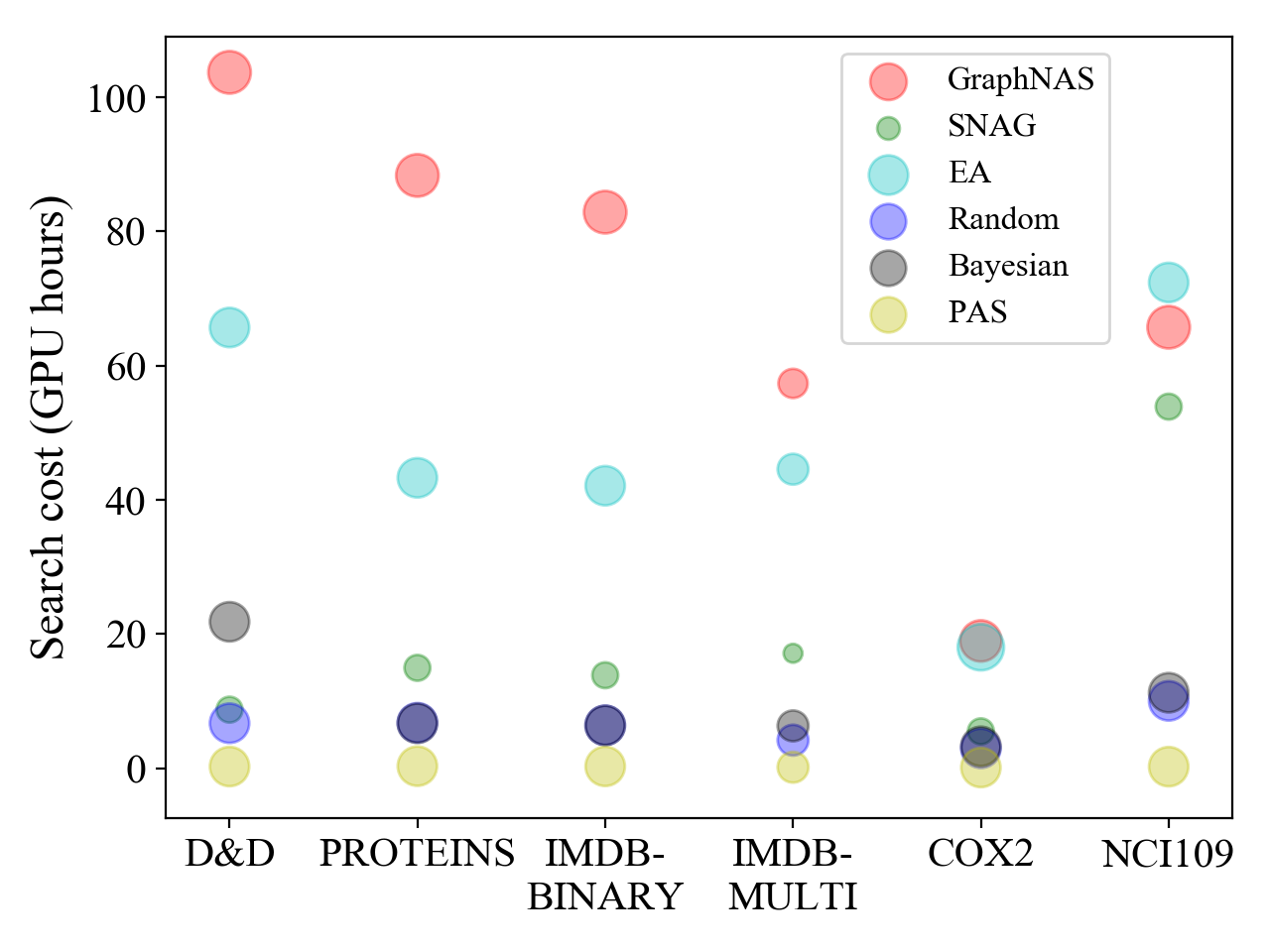}
	%	\vspace{-10pt}
	\caption{(Best viewed in color) The search cost of each model on all datasets. The size of each circle represents the size of the search space each method uses. 
		%		GraphNAS searches for fine-grained parameters in aggregation functions while SNAG only searches for aggregation functions, thus 
		GraphNAS and SNAG have the largest and smallest search space, respectively.
		EA, Random, Bayesian and PAS use the same search space, thus the circles are of the same size.}
	\label{fig-gpuhours}
		\vspace{-10pt}
\end{figure}

%\vspace{-5pt}
\subsection{The Efficiency of PAS}
\label{sec-evaluate-algo}

In this section, we show the efficiency superiority of the differentiable search process of PAS, which relies on the proposed coarsening strategy, over NAS baselines.

%Based on the designed coarsening strategy, PAS has an efficient search process with differentiable search algorithm.
%Here we show the efficiency of PAS compared to NAS baselines.

%\begin{figure}[t]
%	\subfigure[D\&D]{
%		\includegraphics[width=0.45\linewidth]{./fig/dd_efficient}
%	}
%%	\subfigure[PROTEINS]{
%%		\includegraphics[width=0.45\linewidth]{./fig/proteins_efficient}
%%	}
%%	\subfigure[IMDB-BINARY]{
%%		\includegraphics[width=0.45\linewidth]{./fig/imdbb_efficient}
%%	}
%	\subfigure[IMDB-MULTI]{
%	\includegraphics[width=0.45\linewidth]{./fig/imdbm_efficient}
%	}
%%	\subfigure[COX2]{
%%		\includegraphics[width=0.45\linewidth]{./fig/cox2_efficient}
%%	}
%%	\subfigure[NCI109]{
%%		\includegraphics[width=0.45\linewidth]{./fig/nci109_efficient}
%%	}
%	\caption{(Best viewed in color) The validation accuracy of each child net w.r.t. search time (in seconds) in log base 10.}
%	\label{fig-search-efficiency}
%\end{figure}

%Firstly, we compare the search cost of all methods. As shown in Figure~\ref{fig-gpuhours}, the search cost of PAS is the smallest, which is mainly attributed to the designed differentiable search algorithm and coarsening strategy. Compared to Random, Bayesian, and EA, PAS use the same search space. And compared to SNAG, whose search space is smaller than that of PAS, however, the search cost of PAS is still smaller since SNAG is using RL based search algorithm. Thus, we can see the superiority of the coarsening strategy in obtaining an efficient search process.

As shown in Figure~\ref{fig-gpuhours}, we compare the search cost of all NAS methods.
The search cost of PAS is the smallest among all NAS baselines, which is mainly attributed to the designed differentiable search algorithm, thus the coarsening strategy.

Compared with GraphNAS and SNAG, which focus on searching for aggregation functions and directly use global pooling methods, the search space of PAS is more expressive as analyzed in Section~\ref{sec-ablation-pool} and has a moderate size. Besides, with the designed coarsening strategy, PAS can be optimized with gradient descent and achieve the two orders of magnitude reduction of search cost. On the contrast, these RL based methods need thousands of evaluations which are inefficient in particular.
%the advantages of PAS lies not only in the expressive but moderate size search space in learning data-specific pooling architectures, but also in the reduction of times in two orders of magnitude by the designed coarsening strategy. 
%
%
%reduce the search time by two orders of magnititude due to the designed coarsening strategy.
%have a moderate search space size and learn to design data-specific pooling operations in GNNs. 
%PAS is superior to these 2 baselines on data-specific pooling architecture learning due to the expressive search space and the efficient search process brought by the designed coarsening strategy.

Besides, from Figure \ref{fig-gpuhours}, we can see that based on the proposed search space in Section \ref{sec-search-space}, EA, Random, and Bayesian are much slower than PAS, especially on D\&D and NCI109, the size or the number of the graphs are much larger than others, the efficiency gain of PAS is much larger. 
Combining with the SOTA performance in Table~\ref{tb-performance-comparisons}, 
PAS is efficient and effective in learning data-specific pooling architectures, which indicates that the coarsening strategy not only brings efficiency improvement, 
but also the performance gain.

To summarize, based on the unified framework and coarsening strategy, 
we design an effective search space and an efficient differentiable search algorithm, which can learn data-specific pooling architectures for graph classification.
%Coarsening strategy is proposed to address the efficiency problem in learning adaptive pooling architectures.
%The SOTA performance and the least search cost can demonstrate the efficiency and effectiveness of this search algorithm.
% <<<<<<< HEAD
%\footnote{+qm+ in above,
%	we have claimed that ``Coarsening strategy'' is very important.
%	thus, 
%	it is more important to have a comparison
%	of PAS v.s. PAS (without Coarsening strategy).\huancheck{}}
%As shown in Figure~\ref{fig-search-efficiency}, PAS can find a better architecture more quickly, demonstrating the efficiency superiority over these baseline methods.
% =======
% To further show the efficiency of PAS, we show the time cost and the child net validation accuracy in the search phase. For PAS, we extract the architecture as the child net by preserving each operation with the largest weight in the supernet in each epoch, and report the accuracy on validation data. For other methods, we directly use the validation accuracy of each sampled architecture as the child net in each epoch. 
% =======
% >>>>>>> fe265d0052d4f61c254932fb53130d9c83bc0ade
%\footnote{+qm+ in above,
%	we have claimed that ``Coarsening strategy'' is very important.
%	thus, 
%	it is more important to have a comparison
%	of PAS v.s. PAS (without Coarsening strategy).\huancheck{}}

%\vspace{-10pt}
%\begin{figure}[ht]
%	
%	\centering
%	\scriptsize
%	\includegraphics[width=0.9\linewidth]{./fig/pooling_analysis_rename}
%	%	\vspace{-10pt}
%	\caption{Rank analysis of the operations in pooling and readout layers. Lower is better.}
%	\label{fig-graphgym-results}
%\end{figure}
\vspace{-10pt}
\subsection{More Results of GraphGym}
\label{sec-exp-graphgym}

GraphGym~\cite{you2020design} was proposed to evaluate the design dimensions of GNN models, like aggregation functions, number of layers, etc. However, for the graph classification task, GraphGym only uses a fixed global pooling function to generate the graph representations based on node aggregation operations. It cannot evaluate different pooling methods for graph classification.

\begin{figure}[ht]
	\centering
	\scriptsize
	\includegraphics[width=0.5\linewidth]{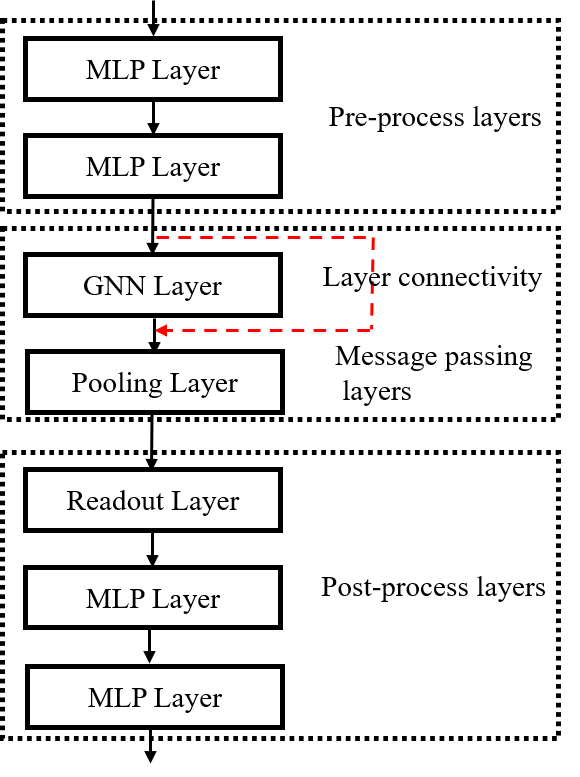}
		\vspace{-10pt}
	\caption{The new search space for graph classification task based on GraphGym.}
	\label{fig-graphgym-search-space}
	\vspace{-15pt}
\end{figure}

To further evaluate the existing pooling methods, we add pooling and readout layers for graph classification on top of GraphGym as shown in Figure~\ref{fig-graphgym-search-space}.
%As shown in Figure~\ref{fig-graphgym-search-space}, we add one pooling layer behind the GNN layer, and add one Readout layer in the post-process stage.
The four candidate operations in pooling layer which denoted as Global, ASAP, SAGPool and Graph U-Net in Figure~\ref{fig-graphgym-winning-ratio}, corresponding to the \texttt{NONE}, \texttt{ASAP}, \texttt{SAGPOOL} and \texttt{TOPKPOOL} in the proposed search space, respectively. Due to the space limit, more experimental details of GraphGym are given in Appendix.

The results are shown in Figure~\ref{fig-graphgym-results}, the upper ones show the average test accuracy rank of each operation on these 420 setups and the bottom ones show the distribution of the accuracy ranking. 

In pooling layer, \texttt{ASAP} and \texttt{SAGPOOL} have a lower average rank and a lower probability to rank last than \texttt{NONE} operation. Combine with the Figure~\ref{fig-graphgym-winning-ratio}, none of these architectures can outperform the other methods, hierarchical pooling methods have advantages over global methods in general. The left part of Figure~\ref{fig-graphgym-winning-ratio} is created by these results.
In readout layer, we use 4 global pooling functions: \texttt{GLOBAL\_MEAN}, \texttt{GLOBAL\_MAX}, \texttt{GLOBAL\_SUM} and  \texttt{GLOBAL\_ATT}. \texttt{GLOBAL\_SUM} have a slightly lower average rank over the other 3 functions, \texttt{GLOBAL\_ATT} have a lower probability to rank last, which means there exist no general global pooling functions that can perform well on various datasets and GNN settings. 

Taking into consideration these experimental results from Figure~\ref{fig-graphgym-winning-ratio} and \ref{fig-graphgym-results}, it shows the need for finding the data-specific pooling architectures for graph classification, which motivates the proposal of PAS in this work.
%	\vspace{-20pt}
\begin{figure}[ht]
	
	\centering
	\scriptsize
	\includegraphics[width=0.8\linewidth]{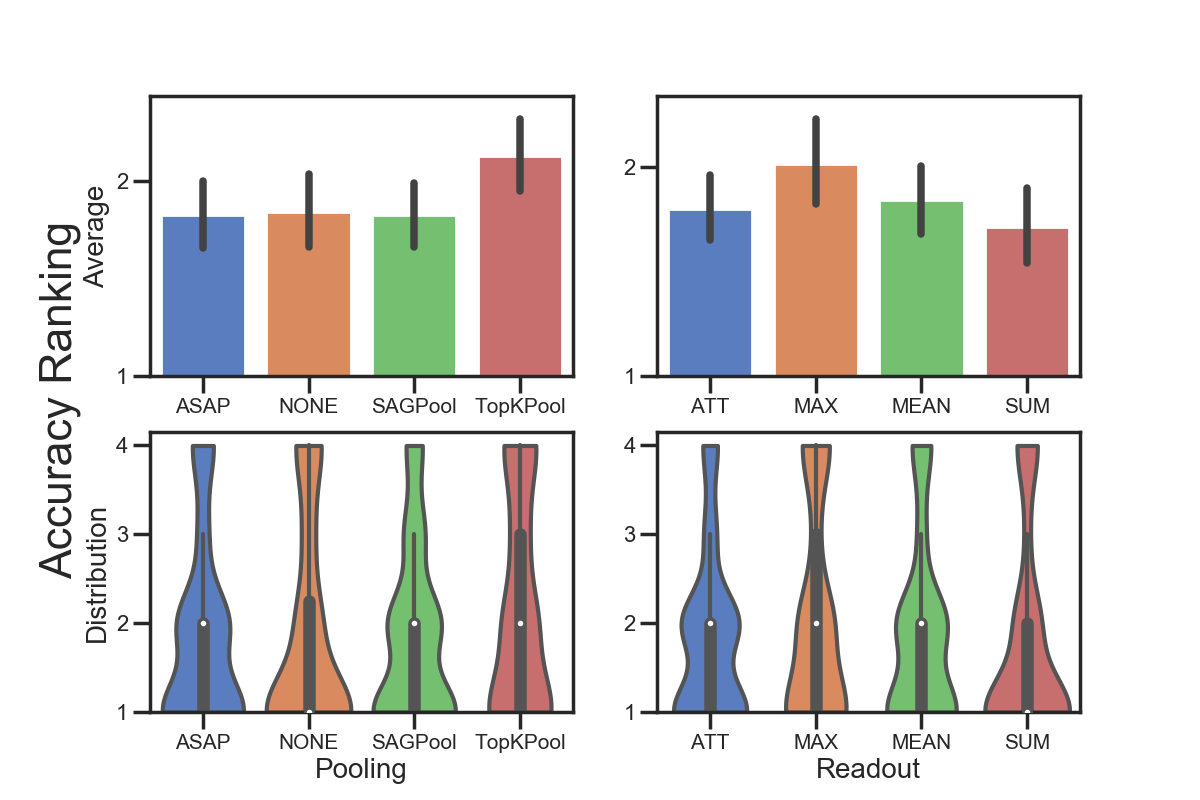}
		\vspace{-10pt}
	\caption{Rank analysis of the operations in pooling and readout layers. Lower is better.}
	\label{fig-graphgym-results}
	\vspace{-10pt}
\end{figure}

%\vspace{-10pt}
\section{Conclusion}
In this paper, we propose a novel framework PAS to automatically learn data-specific pooling architectures for graph classification task. 
By revisiting various human-designed pooling architectures for graph classification, we design a unified framework consisting of four essential modules for graph classification.
Based on this framework, an effective search space is designed by incorporating popular operations from existing human-designed architectures. 
To enable efficient architecture search, we develop a coarsening strategy to continuously relax the search space, thus a differentiable search method can be adopted.
To demonstrate the effectiveness of PAS for graph classification, we conduct extensive experiments on six datasets from three domains.
The experimental results show that PAS can not only search SOTA data-specific pooling architectures for graph classification, but also performs very efficiently than NAS baselines.

For future work, we plan to investigate in depth the connections between the graph properties and the learned pooling architectures, which can help better understanding the graph classification task. Besides, we plan to further evaluate PAS on large-scale datasets, e.g., OGB benchmark \cite{hu2020ogb}.

%\section*{Appendix}
%Due to the space limit, the appendix is provided in the link: \url{https://github.com/AutoML-Research/PAS/blob/main/appendix.pdf}
%\appendix

\clearpage
%\vspace{-10pt}
\bibliographystyle{ACM-Reference-Format}
\balance
\bibliography{ijcai21}
\clearpage
\appendix
\section{Details of Experiment Settings}
\label{sec-appendix-exp}
\subsection{The implementation details of PAS}
All models are implemented with Pytorch~\cite{paszke2019pytorch} on a GPU 2080Ti (Memory: 12GB, Cuda version: 10.2).
Thus, for consistent comparisons of baseline models, we use the implementation of all GNN baselines by the popular GNN library: Pytorch Geometric (PyG) (version 1.6.1)~\cite{Fey/Lenssen/2019}, which provides a unifying code framework ~\footnote{https://github.com/rusty1s/pytorch\_geometric/tree/master/benchmark/kernel} 
for various GNN models. 
Further, we adopt the same data-preprocessing manner by PyG\footnote{https://github.com/rusty1s/pytorch\_geometric/blob/master/benchmark/kernel/datasets.py} and split data by means of a stratification technique with the same seed.

%All the global and hierarchical baselines are tuned individually with hyperparameters like embedding size, learning rate, dropout, etc. 
%%
%For NAS baselines and PAS, we set search and finetune stages to get the 10-fold cross-validation test accuracy.
%In the search stage, the dataset is split into 80\% for training, 10\% for validation and test with the same seed.
%For EA baseline, we follow the experiments in SPOS~\cite{guo2019single}\footnote{https://github.com/megvii-model/SinglePathOneShot} and set the search epoch to 40. For other baselines, the search epoch is 200. 
%%
%In the finetune stage, we tune these searched architectures with same hyperparameters space on Hyperopt~\footnote{https://github.com/hyperopt/hyperopt} based on 10-fold data, and select the final hyperparameters with mean validation accuracy.
%We perform 10-fold cross-validation to evaluate the model performance based on the searched hyperparameters and report the averaged test accuracy and the standard deviations over 10 folds. 

%\subsection{baselines?}
For all human-designed global and hierarchical pooling baselines, we search the layer numbers of this method, global pooling functions, embedding size, dropout rate, and learning rate as shown in Table~\ref{tb-baseline-hypers}. 
Following the DiffPool~\cite{ying2018hierarchical}, we set the pooling rate $k=\frac{L}{10}\times N$ in Eq.~\eqref{eq-s} for all pooling operations where $L$ is the layer number of this method and $N$ is the node number. 
For each model, select 30  hyperparameters settings with Hyperopt~\footnote{https://github.com/hyperopt/hyperopt}, and evaluate each setting on 10-fold cross-validation data. We select settings based on mean validation accuracy then report the final test accuracy and the standard deviations. Besides, we also provide the comparisons of baseline reported results in Table~\ref{tb-reproduce} to show the influence of different settings. 

%We search the hidden size for each globalZ and hierarchical methods, considering that 

We set training and finetuning stages to get the 10-fold cross-validation test accuracy for all NAS baselines and PAS in this paper. In the training stage, the dataset is split into 80\% for training, 10\% for validation and test with the same seed, and we select architectures from the supernet as shown in Alg. \ref{alg-pas}. In the finetuning stage, we tune these searched architectures over the pre-defined space as shown in Table~\ref{tb-nas-hypers} with Hyperopt based on 10-fold data, and select the final candidate with mean validation accuracy.

For all RL-based methods, GraphNAS~\footnote{https://github.com/GraphNAS/GraphNAS}, GraphNAS-WS, SNAG~\footnote{https://github.com/AutoML-Research/SNAG} and SNAG-WS, we set the training epoch to 200. In each training epoch, we sample 10 architectures and use the validation accuracy to update the controller parameters. After training finished, we sample 5 candidates with the controller. 

For EA baseline, we follow the experiments in~\cite{guo2019single}\footnote{https://github.com/megvii-model/SinglePathOneShot}. We set the population size to 50 and the training epoch to 40. In each training epoch, random select an architecture and mutates all operations with probability 0.1 to generate the new architecture; random select two architectures and crossed to generate one new architecture, 25 mutation operations and 25 crossover operations in each training epoch. The architecture is derived based on the validation accuracy after the training stage terminates.

For Random and Bayesian methods, we set the training epoch to 200. In each training epoch, sample one architecture and train from scratch. After training finished, we select one candidate with the validation accuracy.

For PAS, we set the training epoch to 200 as shown in Alg. \ref{alg-pas}. In each training epoch, PAS samples a set of  minibatchs and uses the training loss to update parameters $\textbf{W}$ and use the validation loss to update $\bm{\alpha}$. After search process is finished, we derive the candidate architecture from the supernet. Repeat 5 times with different seeds, we can get 5 candidates.

In the finetuning stage, each candidate architecture owns 30 hyper steps. In each hyper step, a set of hyperparameters will be sampled from Table~\ref{tb-nas-hypers} based on Hyperopt, then we generate final performance on 10-fold data. We choose the hyperparameters for each candidate with the mean validation accuracy. After that, we choose the candidate with the mean validation accuracy then report the final test accuracy and the standard deviations based on 10-fold cross-validation data. 

\begin{table}[ht]
	\small
	\caption{Hyperparameter space for human-designed baselines.}
	\label{tb-baseline-hypers}
	\vspace{-10pt}
	\begin{tabular}{l|l}
		\toprule
		Dimension             & Operation              \\ \midrule
		Layer number     & 1, 2, 3, 4, 5           \\  \hline
		Global pooling function &  \texttt{GLOBAL\_MEAN}, \texttt{GLOBAL\_SUM} \\ \hline
		Embedding size      & 8, 16, 32, 64, 128, 256, 512 \\ \hline
		Dropout rate     & 0, 0.1, 0.2,...,0.9    \\  \hline
		Learning rate    & $[0.001, 0.025]$   \\ 
		\bottomrule
	\end{tabular}
	
\end{table}

\begin{table}[ht]
	
	\small
	\caption{The reported results of 3 methods (denoted as Method1) in other methods (denoted as Method2).}
	\label{tb-reproduce}
	\vspace{-10pt}
	\begin{tabular}{c|c|c|c}
	\toprule
		\multicolumn{1}{c|}{Method 1} & Method 2       & D\&D   & PROTEINS \\ \midrule
		\multirow{6}{*}{DiffPool~\cite{ying2018hierarchical}}     
		& FAIR~\cite{errica2019fair}          & 0.7500 & 0.7370   \\ \cline{2-4} 
		& SAGPool~\cite{lee2019self}       & 0.6695 & 0.6820   \\ \cline{2-4} 
		& DiffPool~\cite{ying2018hierarchical}      & 0.8064 & 0.7625   \\ \cline{2-4} 
		& Graph   U-Net~\cite{gao2019graph} & 0.8064 & 0.7625   \\ \cline{2-4} 
		& ASAP~\cite{ranjan2020asap}          & 0.6695 & 0.6820   \\ \cline{2-4} 
		& PAS           & 0.7775 & 0.7355   \\ \midrule
		\multirow{7}{*}{DGCNN~\cite{zhang2018end}}     
		& FAIR~\cite{errica2019fair}          & 0.7660 & 0.7290   \\ \cline{2-4} 
		& DGCNN~\cite{zhang2018end}        & 0.7937 & 0.7554   \\ \cline{2-4} 
		& Graph U-Net~\cite{gao2019graph}& 0.7937 & 0.7626   \\ \cline{2-4} 
		& DiffPool~\cite{ying2018hierarchical}      & 0.7937 & 0.7554   \\ \cline{2-4} 
		& SAGPool~\cite{lee2019self}       & 0.7253 & 0.6672   \\ \cline{2-4} 
		& ASAP~\cite{ranjan2020asap}         & 0.7187 & 0.7391   \\ \cline{2-4} 
		& PAS           & 0.7666 & 0.7357   \\ \midrule
		\multirow{2}{*}{Graph U-Net~\cite{gao2019graph}}  
		& ASAP~\cite{ranjan2020asap}          & 0.7501 & 0.7110   \\ \cline{2-4} 
		& SAGPool~\cite{lee2019self}       & 0.7501 & 0.7110   \\ \midrule
		\multirow{3}{*}{GraphSAGE~\cite{hamilton2017inductive}}    
		& FAIR~\cite{errica2019fair}         & 0.7290 & 0.7300   \\ \cline{2-4} 
		& DiffPool~\cite{ying2018hierarchical}     & 0.7542 & 0.7048   \\ \cline{2-4} 
		& PAS           & 0.7727 & 0.7375   \\ \bottomrule
	\end{tabular}
\end{table}

\begin{table}[ht]
	\small
	\caption{Hyperparameter space in the finetuning stage for NAS methods.}
	\label{tb-nas-hypers}
	\vspace{-10pt}	
	\begin{tabular}{l|l}
		\toprule
		Dimension     & Operation                    \\ \midrule
		Embedding size   & 8, 16, 32, 64, 128, 256 \\ \hline
		Dropout rate  & 0, 0.1, 0.2, $\cdots$, 0.9          \\ \hline
		Learning rate & $[0.001, 0.025]$             \\ \hline
		Optimizer                          & Adam, AdaGrad                \\ \hline
		Activation function                & RELU, ELU                   \\ \bottomrule
	\end{tabular}

\end{table}

%For PAS, we use the coarsening strategy to solve the node mismatch problem in Pooling Module. For computational efficiency, we drop the nodes $idx_{d}$ which were not selected by any pooling operation after Eq.~\eqref{eq-weightsum-pool}. The nodes $idx_{d}$ can be represented as  $\{idx_d|\sum_{i=1}^{\left|\mathcal{O}_p\right|} c_i \mathbbm{1}(p_v^i)<th, \forall v \}$, where $c_i$ is the weights of $i$-th pooling operation, $\mathbbm{1}(p_v^i)=1$ indicate the node $v$ was selected by $i$-th pooling operation and $\mathbbm{1}(p_v^i)=0$ otherwise, and the threshold $th=0.01$ in this paper.
%we denote the selected node $v$ in $i$-th pooling operation as $\mathbbm{1}(p_v^i)=1$,
%$\{idx_{d}|\mathbbm{1}(<th)=1 \forall v\}$
%%we use the same pooling rate $k=\frac{L}{10}$ in Eq.~\eqref{eq-s}. For computational efficiency, 
%we drop the nodes which was not selected by any pooling operation behind the Pooling Module, whcih can enoted as $\{idx|\mathbbm{1}(>th)=1 \forall v\}$ and threshold $th=0.001$ in this paper.

%\subsection{The Evaluation of different Pooling operations on GraphGym}

\begin{table}[ht]
	\scriptsize
	\centering
	\caption{The search space we use in GraphGym. Other parameters remain the same. }
	\vspace{-10pt}
	\begin{tabular}{l|l}
		\toprule
		Pre-process layer      & 1, 2                             \\ \hline
		Message Passing layers & 1, 2, 3, 4, 5, 6                     \\ \hline
		Post-process layers    & 1, 2                             \\ \hline
		Pooling layer          & \texttt{SAGPOOL}, \texttt{TOPKPOOL}, \texttt{ASAP},  \texttt{NONE} \\   \hline
		Readout layer          & \texttt{GLOBAL\_MEAN}, \texttt{GLOBAL\_MAX}, \texttt{GLOBAL\_SUM}, \texttt{GLOBAL\_ATT}   \\  \hline
		Learning rate          & 0.01                           \\  \hline
		Batch size             & 16                              \\  \hline
		Training epochs        & 100                             \\ \bottomrule
	\end{tabular}
	\label{tb-graphgym-params}
\end{table}

\subsection{The implementation details of GraphGym}
\label{sec-appendix-graphgym}
In this section, we show the details of the designed experiments as mentioned in Figure~\ref{fig-graphgym-winning-ratio}.

%\begin{figure}[ht]
%	\centering
%	\scriptsize
%	\includegraphics[width=0.5\linewidth]{./fig/graphgym_lap}
%	\vspace{-10pt}
%	\caption{The new search space for graph classification task based on GraphGym.}
%		\label{fig-graphgym-search-space}
%\end{figure}

%\begin{figure}[ht]
%
%	\centering
%	\scriptsize
%	\includegraphics[width=0.9\linewidth]{./fig/pooling_analysis_rename}
%\vspace{-10pt}
%	\caption{Rank analysis of the operations in pooling and readout layers. Lower is better.}
%	\label{fig-graphgym-results}
%\end{figure}

Vere recently, GraphGym~\cite{you2020design} was proposed to evaluate the design dimensions of GNN models, like aggregation functions, number of layers, etc. However, for the graph classification task, GraphGym only uses a fixed global pooling function to generate the graph representations based on node aggregation operations. It cannot evaluate different pooling methods for graph classification.
To further evaluate the pooling methods, we design a search space for graph classification on top of GraphGym.
As shown in Figure~\ref{fig-graphgym-search-space}, we add the pooling layer behind the 1-st, 3-rd, 5-th GNN layer, and add one Readout layer in the post-process stage. The search dimensions are shown in Table~\ref{tb-graphgym-params}. 
%The four methods, Global, ASAP, SAGPool and Graph U-Net in Figure~\ref{fig-graphgym-winning-ratio} corresponding to \texttt{NONE}, \texttt{ASAP}, \texttt{SAGPOOL} and \texttt{TOPKPOOL} operations, respectively, in our search space and Table~\ref{tb-graphgym-params}.

We use the 6 real-world datasets and 8 synthetic datasets mentioned in GraphGym.
Each synthetic dataset chooses graph structure function from \{scalefree, smallworld\} and choose feature function from \{clustering, const, onehot, pagerank\}. 256 graphs are generated with different Average Clustering Coefficient and Average Path Length. 
We use 420 setups on these 14 datasets so that each dataset has 30 hits on average, and the results are shown in Figure~\ref{fig-graphgym-winning-ratio} and \ref{fig-graphgym-results}.

\begin{figure}[t]
	\centering
	\includegraphics[width=0.7\linewidth]{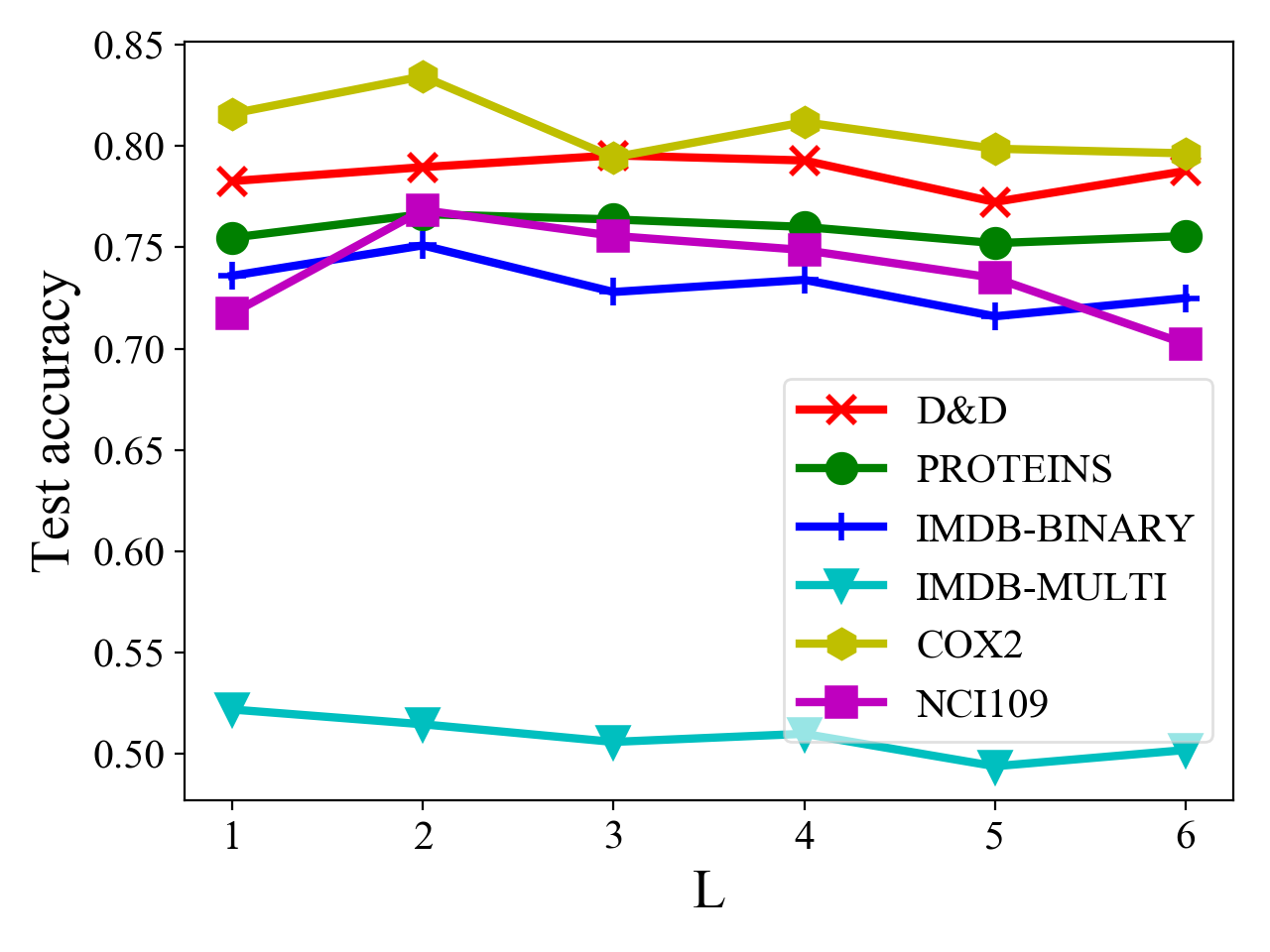}
	\vspace{-10pt}
	\caption{The test accuracy w.r.t. the layer numbers for PAS.}
	\label{fig-layerk}
	\vspace{-10pt}
\end{figure}
\section{Experiments}
\subsection{More figures}
\begin{figure}[ht]
		\subfigure[IMDB-BINARY]{
			\includegraphics[width=0.45\linewidth]{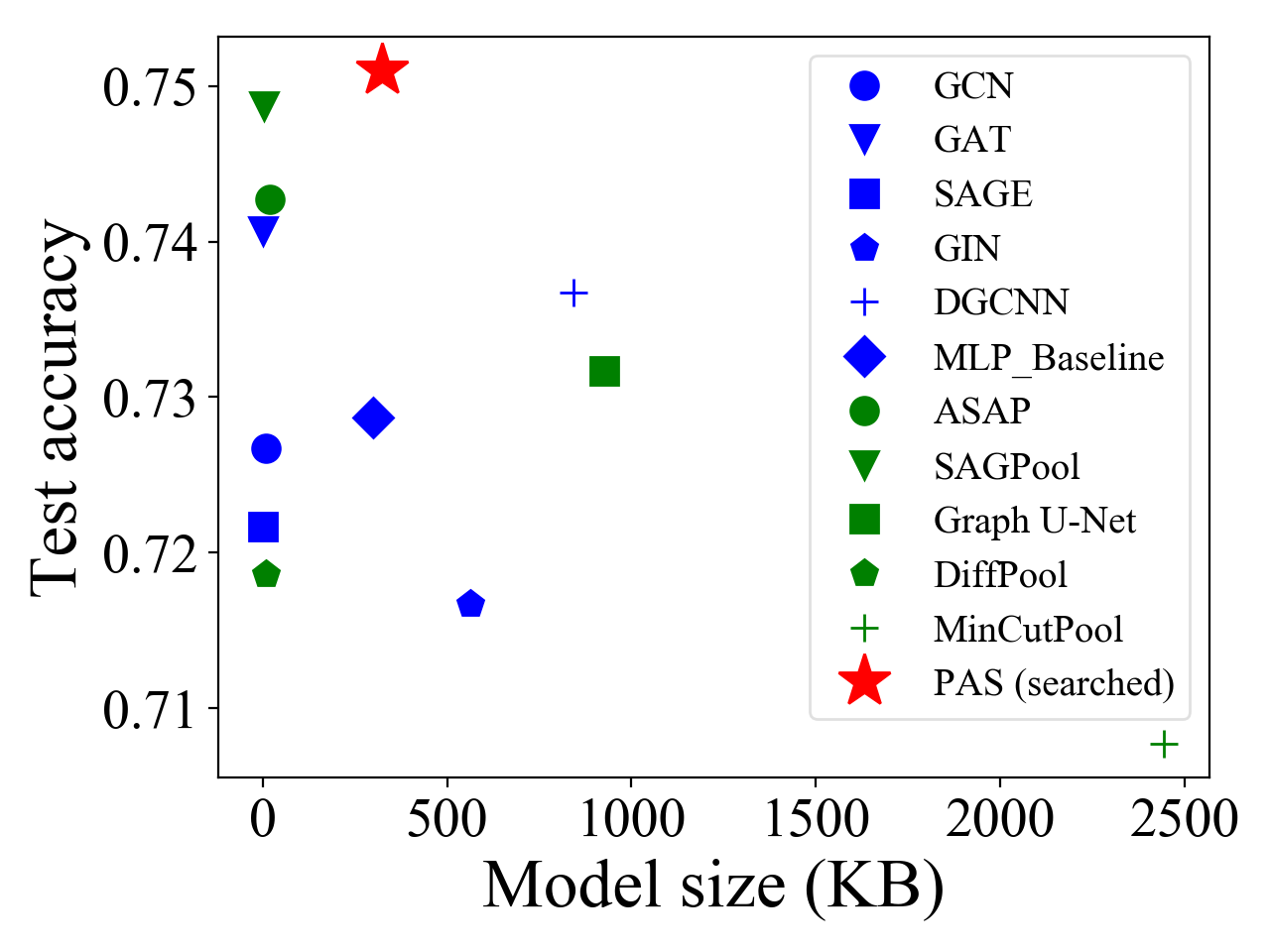}
		}
		\subfigure[IMDB-MULTI]{
		\includegraphics[width=0.45\linewidth]{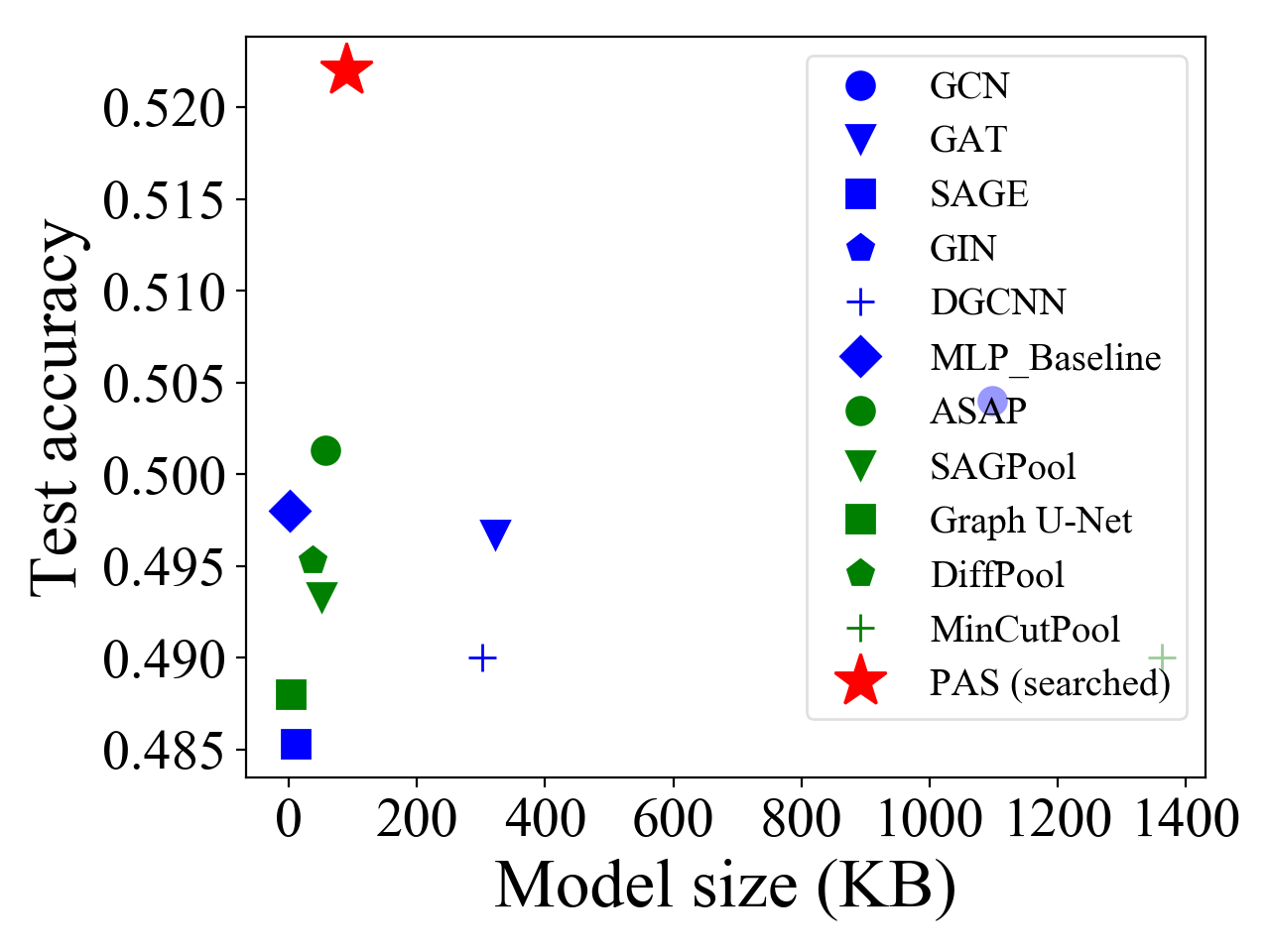}
		}
		\subfigure[COX2]{
		\includegraphics[width=0.45\linewidth]{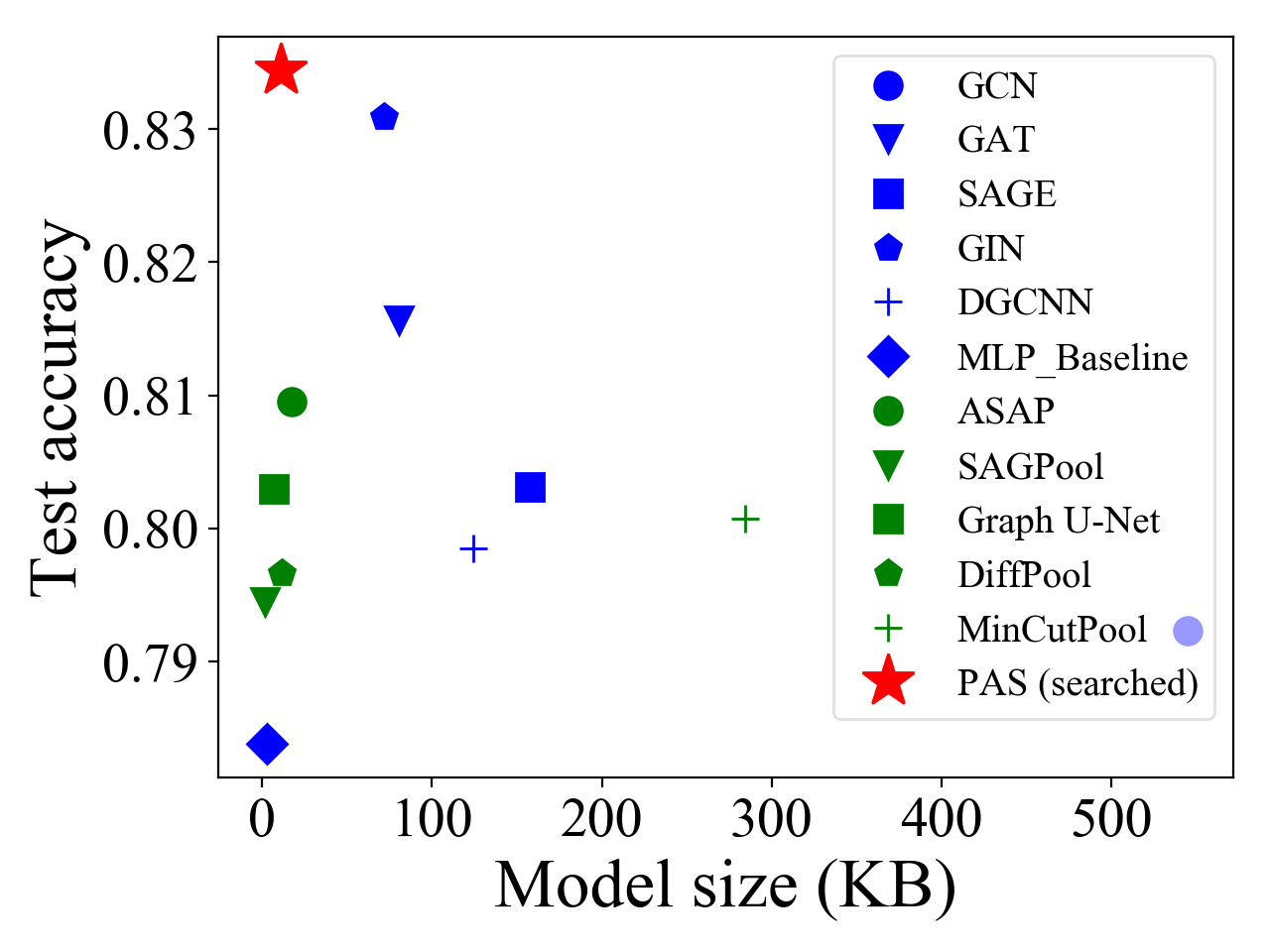}
	}
	\subfigure[NCI109]{
		\includegraphics[width=0.45\linewidth]{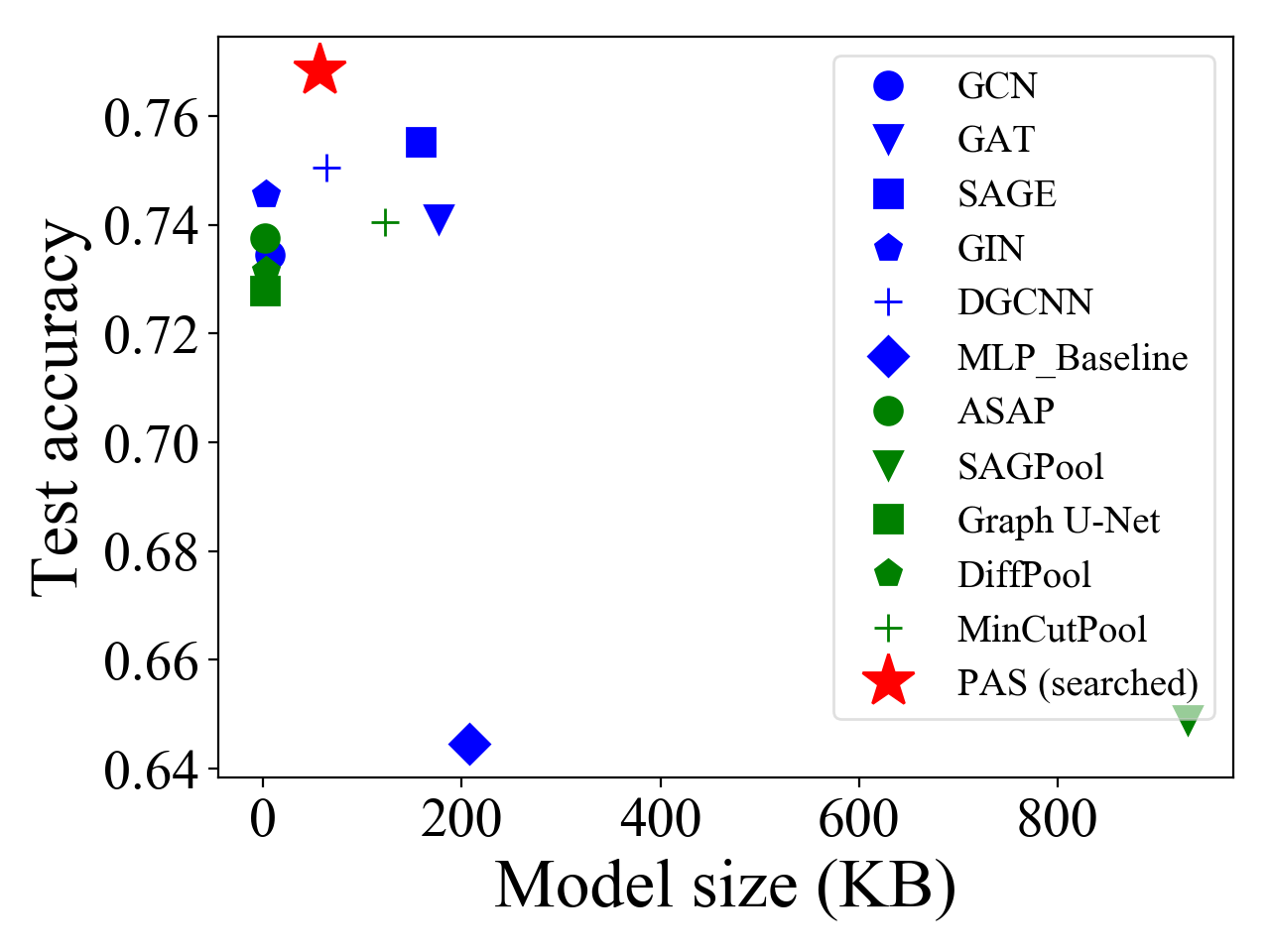}
	}

	\caption{(Best viewed in color) The test accuracy  w.r.t. model size. The searched architectures by PAS can achieve SOTA performance with moderate size in terms of the model parameters.}
	\label{fig-params_acc_all}

\end{figure}

For sake of the space, we only show the test accuracy and model size comparisons on 2 datasets in Figure~\ref{fig-params_acc}. Here, more figures are shown in Figure~\ref{fig-params_acc_all}. 

\subsection{The influence of layers}
\label{sec-appendix-layer}

Here we conduct experiments to show the influences of the layer number $L$ of PAS by varying $L \in \{1,2,3,4,5,6\}$ in PAS.
As introduced in Section~\ref{sec-search-space} and Figure~\ref{fig-framework}(b), each layer consists of one Aggregation and Pooling Module in PAS.
The results are shown in Figure~\ref{fig-layerk}, from which we can see that the trending that the performance increases firstly with the increase of $L$ and then decreases. 
%%\footnote{+qm+ see footnote~\ref{ft:3}.\weicheck{}}
%To be specific, the best performance is obtained when $L=3$ on D\&D, $L=2$ on PROTEINS, IMDB-BINARY, and COX2, and $L=1$ on IMDB-MULTI. 
%They align with the number of layers by PAS on all datasets in Section~\ref{subsec-exp-performance}.\wei{since ``The number of layers on each dataset is by empirical results from the influence of the layer numbers on PAS'', we can not claim it align with Section~\ref{subsec-exp-performance}, reciprocal causation}
Looking back at the statistics of the datasets in Table~\ref{tb-graph-dataset}, we can get an empirical guideline for architecture design for graph classification, although it may not always be true, that the larger number of layers is preferable for graphs in the larger size (more nodes).

\end{document}